\documentclass[review]{elsarticle}
\usepackage{multirow}%
\usepackage{amsmath,amssymb,amsfonts}%
\usepackage{amsthm}%
\usepackage{mathrsfs}%
\usepackage{xcolor}%
\usepackage{textcomp}%
\usepackage{manyfoot}%
\usepackage{booktabs}%
\usepackage{algorithm}%
\usepackage{algorithmicx}%
\usepackage{algpseudocode}%
\usepackage{listings}%
\usepackage{caption}
\usepackage{tabularx} 
\usepackage{amsmath} 
\usepackage{ragged2e} 
\usepackage{caption}
\usepackage{hyperref}
\usepackage{graphicx}
\usepackage{amsmath, amssymb}
\usepackage{threeparttable}
\usepackage{booktabs, multirow, colortbl}
\usepackage[table]{xcolor}
\usepackage{caption}
\usepackage{siunitx}
\usepackage{makecell}
\usepackage{geometry}
\usepackage{tabularx}
\usepackage{ragged2e}
\usepackage{graphicx}
\usepackage{amsmath, amssymb, amsfonts}
\usepackage{url}
\usepackage[title]{appendix}
\usepackage{hyperref}
\usepackage{algorithm}      
\usepackage{algpseudocode}
\usepackage[utf8]{inputenc}
\usepackage{threeparttable}  
\usepackage{lscape}
\usepackage{newtxtext,newtxmath}   
\usepackage{makecell} 
\usepackage{caption}  
\usepackage{graphicx} 
\usepackage{booktabs}
\usepackage{rotating} 
\usepackage[utf8]{inputenc}
\usepackage{adjustbox}
\usepackage{arydshln}
\hypersetup{pdfencoding=unicode,psdextra}
\usepackage{longtable}
\usepackage{natbib}
\usepackage{makecell}   
\usepackage{threeparttable} 
\usepackage{siunitx}
\usepackage{adjustbox}  
\usepackage{amsmath} 
\usepackage{rotating}     
\usepackage{booktabs}
\usepackage{microtype}
\usepackage{geometry}
\usepackage{microtype}
\usepackage{inputenc}
\usepackage{geometry} 
\usepackage[T1]{fontenc}
\usepackage{lmodern}
\usepackage{tabularx}
\newcolumntype{Y}{>{\centering\arraybackslash}X}
\usepackage{caption}      
\usepackage{array}
\newcolumntype{C}{>{\centering\arraybackslash}p{1.2cm}}
\usepackage{pifont}   
\usepackage{array}        
\usepackage[utf8]{inputenc}
\hypersetup{
    colorlinks=true,
    linkcolor=black,
    citecolor=blue,
    urlcolor=blue,
    pdftitle={DiffKD-DCIS: Predicting Upgrade of Ductal Carcinoma In Situ},
    pdfauthor={Tao Li et al.},
    unicode=true
}

\usepackage{lineno,hyperref}
\modulolinenumbers[5]

\journal{Journal of \LaTeX\ Templates}

\begin{document}

\begin{frontmatter}

\title{Conditional Diffusion Model with Anatomical-Dose Dual Constraints for End-to-End Multi-Tumor Dose Prediction}

\author{\texorpdfstring{Hui Xie$^{a,b}$}{Hui Xie (a,b)}}
\author{\texorpdfstring{Haiqin Hu$^{c}$}{Haiqin Hu (c)}}
\author{\texorpdfstring{Lijuan Ding$^{d}$}{Lijuan Ding (d)}}
\author{\texorpdfstring{Qing Li$^{b}$}{Qing Li (b)}}
\author{\texorpdfstring{Yue Sun$^{a,\ast}$}{Yue Sun (a,*)}}  
\author{\texorpdfstring{Tao Tan$^{a,\ast}$}{Tao Tan (a,*) }}

\affiliation{organization={a. Faculty of Applied Sciences, Macao Polytechnic University},
            addressline={R. de Luís Gonzaga Gomes}, 
            city={Macao},
            postcode={999078}, 
            state={},
            country={P. R. China}}
\affiliation{organization={b. Department of Oncology,Affiliated Hospital of Xiangnan University},
            addressline={No.31 Renmin West Road}, 
            city={Chenzhou},
            postcode={423000}, 
            state={Hunan},
            country={P. R. China}}

\affiliation{organization={c. Department of Oncology,Jiangxi Cancer Hospital},
            addressline={No. 519, Beijing East Road}, 
            city={Nanchang},
            postcode={330029}, 
            state={Jiangxi},
            country={P. R. China}}

\affiliation{organization={d. Department of Oncology, Chenzhou Third People's Hospital},
                addressline={No. 8, Jiankang Road}, 
            city={Chenzhou},
            postcode={423000}, 
            state={Hunan},
            country={P. R. China}}

\cortext[mycorrespondingauthor]{Correspondingauthor: Tao Tan and Yue Sun}
\ead{taotanjs@gmail.com and joyyuesun@gmail.com}

\address[mymainaddress]{ }
\address[mysecondaryaddress]{  }

\begin{abstract}
Traditional manual optimization in radiotherapy planning is highly experience-dependent and time-consuming. Although deep learning offers automated solutions, existing models still face limitations in generalization across tumor types, prediction accuracy, and adherence to clinical constraints. This paper proposes a novel end-to-end multi-tumor dose prediction model—the Anatomical-Dose Dual Constraints Diffusion Model (ADDiff-Dose)—which innovatively integrates anatomical and clinical dose dual constraints into a conditional diffusion framework. We design a Lightweight 3D Variational Autoencoder (LightweightVAE3D) that compresses high-resolution CT images to 0.3\% of their original size for efficient 3D data processing. In the latent space, a conditional diffusion model with a 3D U-Net backbone guides the denoising process to generate dose distributions by incorporating multimodal conditions—including target and organ-at-risk masks and beam parameters—via a multi-head attention mechanism. Furthermore, we introduce a composite loss function that not only optimizes the mean squared error of noise prediction but also, for the first time, directly "hard-codes" over 50 clinical dose-volume constraints into the optimization objectives, ensuring the clinical feasibility of the generated results. Extensive evaluations on a large-scale public dataset (2,877 cases) and three external institutional cohorts (450 cases in total) demonstrate that ADDiff-Dose significantly outperforms state-of-the-art models across nearly all key metrics. Qualitative assessments include visual comparisons and difference maps, confirming superior structural dose distributions with reduced artifacts in high-dose regions. Ablation studies further confirm that our proposed anatomical-dose dual-constraint design is core to the model's performance: removing any constraint leads to significant performance degradation, with the clinical constraint loss (\(L_{\text{cond}}\)) ablation increasing spinal cord \(D_{\text{max}}\) error by 109\%, and the MSE loss (\(L_{\text{mse}}\)) ablation reducing spatial accuracy (DICE) by 8.8\%. The full model enhances clinical dose compliance by 28.5\% and enables uncertainty quantification through multiple stochastic inferences.This work provides a highly accurate, efficient, and generalizable automated treatment planning tool for clinical practice.
\end{abstract}

\begin{keyword}
\texttt{Conditional Diffusion Model; Anatomical-Dose Dual Constraints; Dose Prediction; Multi-Tumor}
\MSC[2010] 00-01\sep  99-00  

\(\#\) :These authors contributed equally to this work.

\end{keyword}

\end{frontmatter}


\section{Introduction}
\label{１}

Cancer remains a leading cause of morbidity and mortality worldwide, posing a persistent threat to human health. Among the three principal treatment modalities—surgery, chemotherapy, and radiotherapy—approximately 80\% of patients with malignant tumors require radiotherapy at some stage during their disease course~\cite{mak2019use}. The efficacy of radiotherapy hinges on the precise formulation of treatment plans, which aim to maximize tumor cell eradication while sparing adjacent healthy tissues. This planning process typically demands experienced radiotherapy physicists to iteratively adjust beam parameters to generate clinically acceptable three-dimensional dose distributions, often assessed via dose-volume histograms (DVHs). However, manual planning is time-consuming and labor-intensive, usually taking several hours to days for a single case. Additionally, the final plan quality can vary significantly depending on the planner’s expertise, potentially limiting the individualization and optimality of treatment regimens~\cite{MINNITI2012215}.

In response, recent advances in deep learning have facilitated the development of data-driven models that automate dose prediction and streamline clinical workflows. Convolutional neural networks (CNNs), generative adversarial networks (GANs), and attention-based models have shown promise in learning mappings between anatomical features and dose distributions from historical datasets~\cite{JIANG2024100792}. For example, ResNet101 has been employed to accurately predict dose for nasopharyngeal carcinoma by integrating anatomical and beam parameters with a multi-scale feature fusion mechanism~\cite{chen2021dvhnet} \cite{chen2019feasibility}. In the domain of lung cancer, the asymmetric A-Net architecture achieved clinically acceptable results in the 50–60 Gy range and delivered high spatial precision in mobile lesion regions~\cite{7785132,9204582}. Traditional knowledge-based planning (KBP) approaches also remain relevant. Techniques combining kernel density estimation, k-nearest neighbors(KNN), and principal component analysis (PCA)  have demonstrated the feasibility of dose modeling from limited clinical cohorts~\cite{li2022personalized}.
Despite encouraging progress, current methods face several limitations. First, most existing models are tailored to specific tumor types (e.g., prostate, head and neck) or limited to particular delivery techniques (e.g., IMRT, Intensity-Modulated Radiation Therapy; VMAT, Volumetric-Modulated Arc Therapy), limiting cross-domain generalization~\cite{9204582, ahn2021deep}. Second, training data volumes remain relatively small (typically <200 patients), restricting model robustness to anatomical variability~\cite{Kontaxis_2020}. Third, prediction performance has yet to meet clinical gold standards: even advanced architectures like U-Net and 3D GAN exhibit mean absolute errors exceeding 1–3 Gy in certain regions or cases~\cite{liu2025dose, Gao_2023_CVPR}.

In parallel, diffusion models have emerged as a powerful generative paradigm, achieving state-of-the-art performance across diverse domains such as image synthesis~\cite{shen2025imagharmony}, fashion generation~\cite{shen2025imaggarment}, and human motion modeling~\cite{shen2024advancingposeguidedimagesynthesis, shen2025long}. Their ability to iteratively refine predictions through a learned denoising process offers unique advantages for generating complex, high-resolution outputs under multimodal constraints. However, while diffusion models have been widely adopted in computer vision, natural language processing, and other multimedia applications~\cite{shen2024boostingconsistencystoryvisualization, shen2024imagpose, shen2024imagdressingv1customizablevirtualdressing}, their application to radiotherapy dose prediction remains relatively nascent. Only a handful of recent studies, such as DoseDiff~\cite{10486983} and MD-Dose~\cite{fu2025mddosediffusionmodelbased}, have begun exploring this avenue, highlighting a critical research gap in leveraging the strengths of diffusion models to handle the anatomical complexity and precision demands of radiotherapy. Existing methods still face unresolved limitations: traditional models like U-Net and GAN fail to integrate clinical dose constraints into core optimization, leading to predicted distributions that may violate safety thresholds; diffusion-based approaches such as DoseDiff~\cite{10486983} and MD-Dose~\cite{fu2025mddosediffusionmodelbased} focus solely on anatomical distance or architectural design, lacking synergistic optimization of 'anatomical structure - dose distribution'. 

To address these challenges, we propose ADDiff-Dose, an end-to-end anatomical-dose dual-constrained conditional diffusion framework for multi-tumor radiotherapy dose prediction. Our model incorporates CT images, organ and target contours, prior dose maps, and beam parameters as multimodal inputs to guide a conditional generative process. By integrating domain knowledge directly into the diffusion trajectory, ADDiff-Dose enables accurate and generalizable dose prediction for both head and neck and lung cancers under a unified architecture. Extensive experiments on multicenter datasets demonstrate its superiority in prediction accuracy, structural consistency, and cross-site adaptability, offering a promising solution for future intelligent radiotherapy planning. While pioneering studies such as DoseDiff ~\cite{10486983} and MD-Dose ~\cite{fu2025mddosediffusionmodelbased} have demonstrated the potential of diffusion models for dose prediction, their conditioning mechanisms primarily focus on anatomical distance or Mamba-based architectures, respectively. Our work, ADDiff-Dose, distinguishes itself by introducing a comprehensive anatomical-dose dual-constraint framework. This framework explicitly encodes over 50 clinical dose-volume constraints directly into the diffusion loss function, ensuring the generated dose distributions are not only accurate but also strictly compliant with clinical protocols—a critical aspect not jointly emphasized in prior diffusion-based approaches. To validate this, we provide both quantitative metrics and qualitative visualizations comparing ADDiff-Dose with these diffusion-based baselines, demonstrating improved dose conformity and fewer violations of clinical constraints. The core innovations of this study are condensed into three key aspects:

1. \textbf{Anatomical-dose dual-constraint framework}:A novel conditioning mechanism that fuses multimodal anatomical features 
(tumor/OAR masks, CT intensity) with over 50 explicit clinical dose-volume constraints (e.g., spinal cord $D_{\text{max}} \leq 45\,\text{Gy}$, lung $V_{20} \leq 30\%$), breaking through the limitation of pure anatomical guidance and ensuring both 
spatial accuracy and clinical compliance.

2. \textbf{Efficient 3D data processing}: A Lightweight 3D-VAE specifically designed for radiotherapy scenarios, achieving 99.7\% dimensional compression of high-resolution CT data while preserving key anatomical features, significantly reducing computational burden and enabling clinical deployment.

3. \textbf{Unified multi-tumor prediction architecture}: Breaking through the limitations of traditional single-tumor models, our framework enables unified intensity-modulated radiation therapy (IMRT) and volumetric modulated arc therapy (VMAT) dose prediction for both head-and-neck (H\&N) and lung cancers within a single architecture. Specifically, a single model checkpoint is trained on a combined dataset (Lung + H\&N), wherein tumor type (i.e., class label) is embedded as a conditional feature to mitigate inter-tumor-type interference and enhance task-specific adaptation.

\section{Materials and Methods}

Radiotherapy dose distribution prediction must simultaneously satisfy the dual requirements of anatomical structure fidelity and physical constraint compliance. To address the limitations of traditional unconditional diffusion models in integrating multimodal prior knowledge, this study proposes an end-to-end diffusion model based on anatomical-dose dual constraints. The model innovatively integrates multimodal inputs and prior-knowledge-guided diffusion mechanisms, enabling accurate dose prediction across tumor types within a unified architecture. The model comprises the following core components: a lightweight 3D variational autoencoder (LightweightVAE3D), a 3D UNet backbone network\cite{sun2025generation}, a conditional embedding layer (ConditionalLayer), and a multi-head attention mechanism (MultiHeadAttention)\cite{vaswani2023attentionneed} (Figure 1, 2 and 3). As shown in Figure~1, ADDiff-Dose operates in two stages: (1) The \textit{LightweightVAE3D} is pre-trained to compress high-resolution CT images and dose map into a low-dimensional latent space. (2) The \textit{conditional diffusion model} then operates entirely within this latent space. It takes the encoded latent vector \( z_0 \) and, through a progressive noising and denoising process conditioned on multimodal conditions such as clinical prior (radiotherapy dose OAR constraint table), Targets/OARs mask, Beam Information, and positional information (tumor distribution site), learns to predict the corresponding dose distribution in the latent domain. The final dose map is obtained by decoding the denoised latent vector back to the image space using the VAE decoder. 

\begin{figure}
    \centering
    \includegraphics[width=1\linewidth]{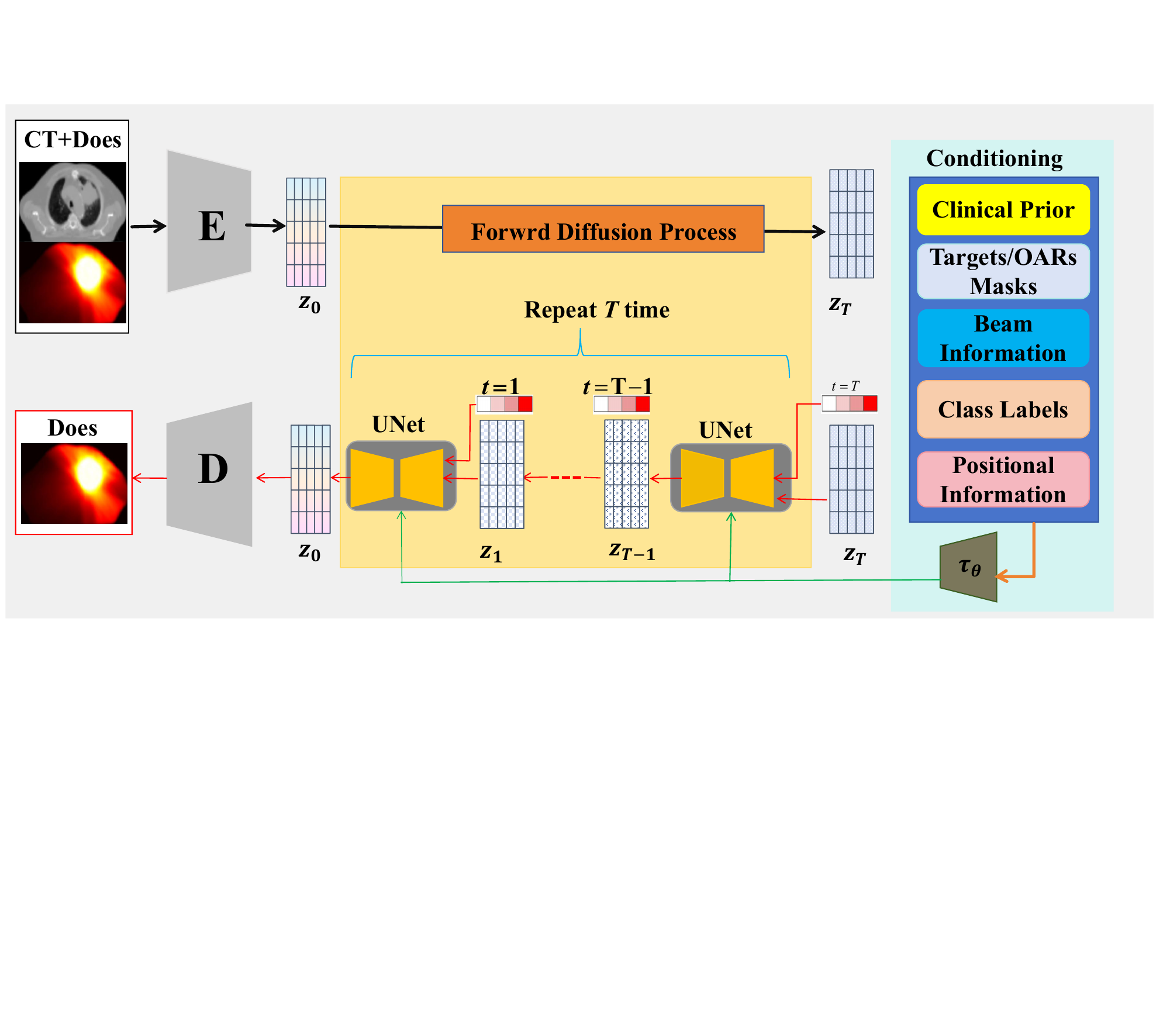}
    \caption{Overall Framework of the Anatomical-Dose Dual Constraints Conditional Diffusion Model (ADDiff-Dose). The schematic illustrates the end-to-end architecture for multi-tumor radiotherapy dose prediction. The diagram illustrates the conditional diffusion process for dose prediction. The top path shows the forward diffusion process, where the initial latent representation \(Z_o\) is progressively noised over T steps to become \(Z_T\). The bottom path shows the reverse denoising process, where a U-Net iteratively refines\(Z_T\) back to  \(Z_o\), conditioned on multi-modal inputs including clinical priors, PTV/OAR masks, class labels, positional information, and beam parameters. The final predicted dose is obtained by decoding the denoised \(Z_o\).}
    \label{fig:placeholder}
\end{figure}

\subsection{Lightweight 3D Variational Autoencoder (LightweightVAE3D)}
\label{subsec2.1}

To address the issue of dramatically increased computational complexity caused by the high resolution of the original CT images in radiotherapy dose prediction tasks, this study designed a lightweight 3D variational autoencoder (LightweightVAE3D)\cite{foti20223dshapevariationalautoencoder}. The model adopts a symmetric encoder-decoder architecture (Figure 2), and achieves a balance between computational efficiency and anatomical feature preservation through joint optimization of feature compression and reconstruction.

\textbf{Encoder Design}:  
The encoder employs a four-layer 3D convolutional chain (kernel size: 4×4×4, stride: 2, padding: 1, activation function: StableSiLU) to perform progressive downsampling. The number of input channels increases gradually from 1 to 256 (channel expansion ratio: 1:256), enabling the model to capture hierarchical image information through multi-scale feature extraction. Ultimately, the original CT images (with a resolution of 96×128×144 voxels) are compressed into a low-dimensional latent space of 6×8×9 voxels, achieving a dimensionality reduction of approximately 99.7\(\%\). At the end of the encoder, fully connected layers output the Gaussian distribution parameters — mean \((\mu)\) and variance \((\sigma^2)\) — of the latent space. These parameters provide the probabilistic foundation for the subsequent diffusion process.

\textbf{Decoder design}: The decoder employs a symmetric 3D transposed convolution chain to gradually upsample and restore image resolution. The number of output channels decreases from 256 to 1, adapting to the normalized CT value range of $[0,1]$. A Sigmoid activation function is introduced to constrain the reconstructed value range, avoiding numerical overflow issues and ensuring the accuracy of anatomical structure reconstruction and clinical rationality.

Training optimization strategy: During the training phase, reparameterization techniques are introduced to enhance gradient propagation efficiency. The model jointly optimizes the adversarial loss and the reconstruction loss $(L_{VAE} = \left\lVert \tilde{x} - x \right\rVert_1 + \beta D_{KL}(q(z|x)\|p(z)))$. Here, the Kullback-Leibler (KL) divergence $D_{KL}$ adjusts the weight coefficient $\beta$ through a dynamic annealing strategy, achieving progressive regularization of the latent space distribution. This strategy reduces computational load while successfully preserving the morphological features of the target area and organs at risk, providing high-quality anatomical priors for subsequent diffusion models.

\begin{figure}
    \centering
    \includegraphics[width=1\linewidth]{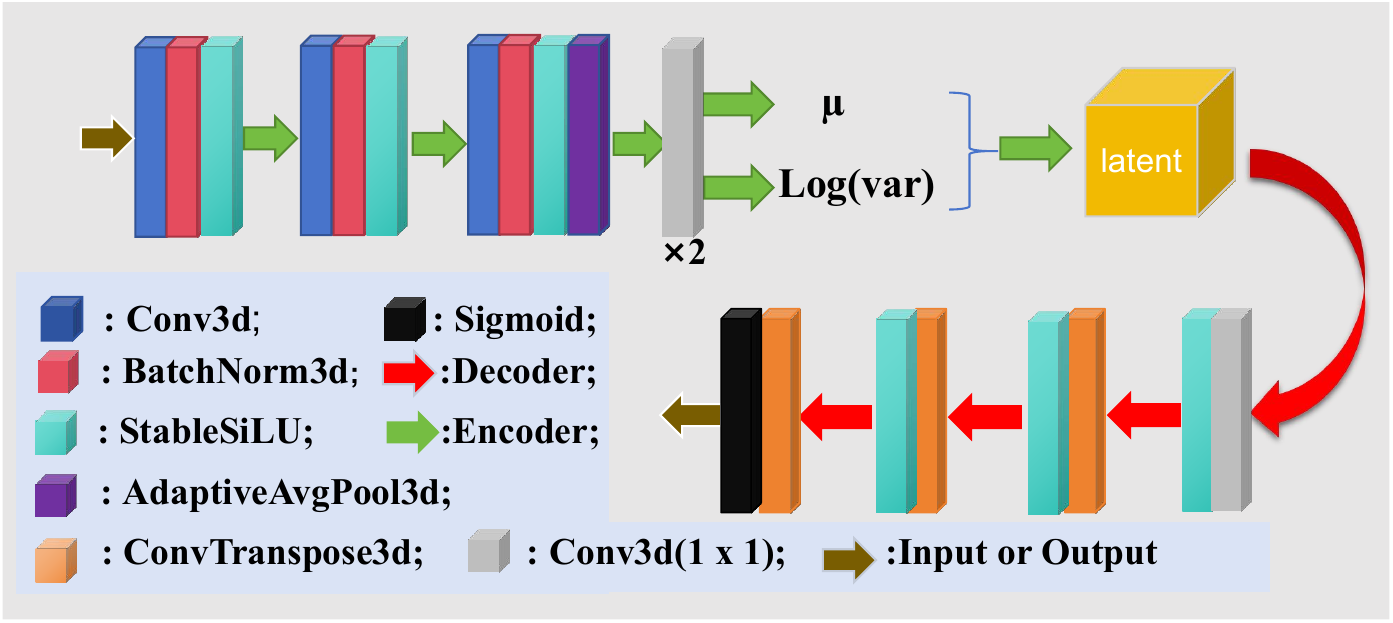}
    \caption{ Architecture Diagram of the Variational Autoencoder (VAE). The upper encoding process uses layers like Conv3d, BatchNorm3d, StableSiLU, and AdaptiveAvgPool3d to generate \(\boldsymbol{\mu}\) and Log(var) for sampling the latent representation. The lower decoding process leverages ConvTranspose3d, StableSiLU, and Sigmoid layers to reconstruct output from the latent space, illustrating the full VAE workflow. }
    \label{Architecture Diagram of the Variational Autoencoder (VAE)}
\end{figure}

\subsection{Conditional Diffusion Model}
\label{subsec2.2}

\subsubsection{Multi-Source Conditional Feature Construction}

After being processed by the VAE encoder, the 3D CT image generates a 32-dimensional latent vector \(z_0\), which serves as the initial input for the diffusion process. Simultaneously, multi-source conditional features are constructed as follows:

\begin{enumerate}
    \item \textbf{Structural features:} Channel merging operations are performed on the planning target volume (PTV) mask and the organs-at-risk (OAR) masks (e.g., spinal cord, lung, brainstem) to generate a multi-channel structural tensor, which is then projected via 3D convolution to form structural conditional features.
    
    \item \textbf{Clinical prior features:} Encode over 50 clinical dose-volume constraints (detailed in Supplementary Table 1), such as spinal cord $D_{\text{max}} \leq 45\,\text{Gy}$, lung $V_{20} \leq 30\%$, and PTV $D_{95} \geq 95\%$ of prescribed dose, into feature vectors through fully connected layers.
    
    \item \textbf{Beam parameters:} Include beam energy (6\,MV/15\,MV), field angle, and number of beams as discrete features, which are converted into continuous vectors via embedding layers.
    
    \item \textbf{Temporal features:} For discrete time steps $t \in [0, 999]$, map them into high-dimensional temporal feature vectors $t_{\text{emb}}$ through a time embedding function to capture the dynamic evolution characteristics during the diffusion process.
\end{enumerate}

All conditional features are fused via the multi-head attention mechanism to form the final conditional feature $\mathbf{C}$, which guides the denoising process of the diffusion model.

\subsubsection{Progressive Noising and Denoising Mechanism}

The noising process follows the Markov chain rule\cite{10.1093/jrsssb/qkae005}, using a linearly scheduled noise coefficient $\beta_t$ (whose values range from $10^{-4}$ to 0.02), progressively injecting Gaussian noise into the latent vector:

\begin{equation}
    z_t = \sqrt{\alpha_t} z_{t-1} + \sqrt{1 - \alpha_t} \varepsilon_{t-1},
    \label{eq:diffusion_forward}
\end{equation}
where $\alpha_t = 1 - \beta_t$, $\varepsilon_{t-1} \sim \mathcal{N}(0, \mathbf{I})$. Through this process, the original signal $z_0$ gradually degrades to Gaussian noise $z_T$.

The reverse denoising process is implemented by a 3D UNet (UNet3D) (Figure 3). This UNet adopts an encoder-decoder architecture, consisting of 4 layers of downsampling (channel numbers increasing from 64 to 512) and 4 layers of upsampling (channel numbers decreasing from 512 to 64), and fuses multi-scale features through skip connections. In the deeper layers, a multi-head attention mechanism is introduced.

The UNet takes the noisy vector $Z_t$, time embedding $\text{temb}$, and conditional feature $C$ as inputs, predicting the noise $\epsilon_\theta(z_t, t, C)$, and then recovers the true signal through the recursive formula:

\begin{equation}
    z_{t-1} = \frac{1}{\sqrt{\alpha_t}} \left( z_t - \frac{1-\alpha_t}{\sqrt{1-\alpha_t^\text{cum}}} \varepsilon_\theta(z_t, t) \right) + \sqrt{\beta_t'} \varepsilon,
    \tag{2}  
    \label{eq:reverse_step}
\end{equation}

where $\alpha_t^\text{cum}$ is the cumulative noise coefficient, and $\beta_t'$ is the adjusted noise variance, used to recover the true signal.

To leverage the stochastic nature of diffusion models, we perform multiple inferences (e.g., 10 runs) on the same input by sampling different noise seeds in the reverse process. This generates a distribution of possible dose maps, from which we compute pixel-wise variance as an uncertainty map, highlighting regions of high variability (e.g., near tumor boundaries).

\begin{figure}[t]
    \centering
    \includegraphics[width=1\linewidth]{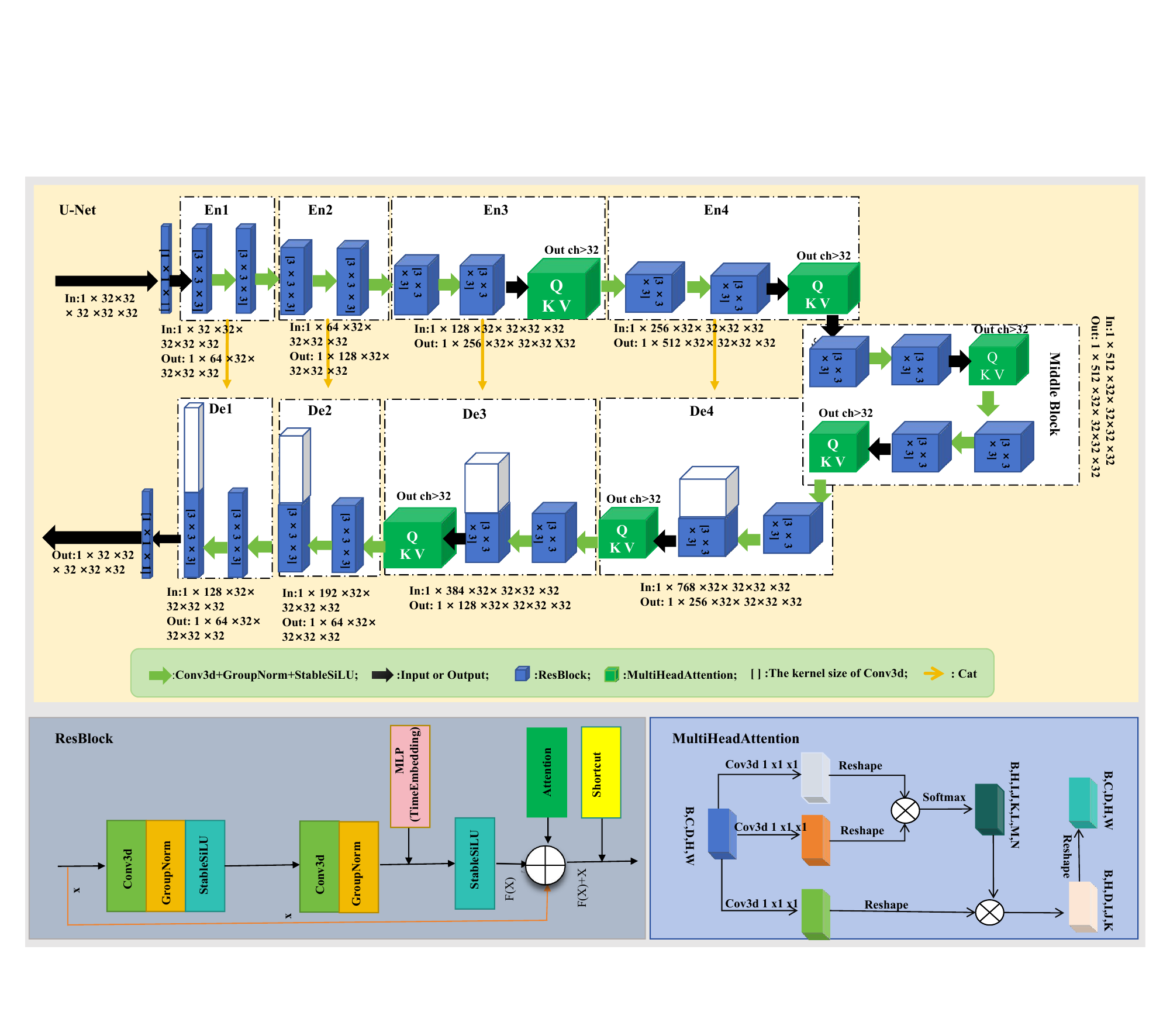}
    \caption{Architecture Diagram of the U-Net Model with Modified Attention and ResBlock Components. It shows the encoding (En1-En4) and decoding (De1-De4) paths, incorporating elements like Cascade - GroupNorm - StableSiLU, ResBlock, and MultiHeadAttention, along with detailed internal structures of ResBlock and MultiHeadAttention at the bottom.}
    \label{Architecture Diagram of the U-Net Model with Modified Attention and ResBlock Components}
\end{figure}
\subsubsection{Anatomical Condition Fusion Mechanism  } 

The conditional feature $C$ and the time embedding $t_{emb}$ interact through the conditional embedding layer. First, the dimensionality of the structural tensor is compressed using 1×1 convolution. Then, cross-attention calculation is performed through Transformer blocks\cite{he2024simplifyingtransformerblocks}. Finally, feature fusion is achieved with the aid of the multi-head attention mechanism. This mechanism ensures that the diffusion process is fully aware of the distribution of anatomical structures, thereby guiding the dose distribution to comply with clinical constraints, such as the maximum dose to the Spinal cord $D_{max} \leq 45$ Gy, the Lung $V_{20} \leq 30\%$, etc. ( Supplementary Table 1).

\subsection{Loss Function}

The composite loss function designed in this study consists of three core components, which are optimized through multi-component collaborative optimization mechanisms to ensure that the generated radiotherapy dose distribution is consistent with the anatomical structure while strictly satisfying clinical compliance requirements. The formula is expressed as:

\begin{equation}
    L = \lambda_1 \cdot L_{\text{mse}} + \lambda_2 \cdot L_{\text{cond}} + \lambda_3 \cdot L_{\text{kl}},
    \tag{3}  
    \label{eq:total_loss}
\end{equation}

1. \textbf{Reconstruction Loss ($L_{mse}$) : }
The reconstruction loss $L_{mse}$ measures the difference between the predicted noise by UNet3D and the actual added noise, thereby driving the learning process of the "noise-to-dose" mapping. It is defined as:

\begin{equation}
    L_{mse} = E_{t, z, C} \left[ \| \epsilon_\theta(z_t, t, C) - \epsilon \|^2 \right].
     \tag{4}
    \label{eq:total_loss11}
\end{equation}

Where:
$\boldsymbol{z}_t$: The latent vector at time step $t$ (generated by 3D-VAE encoding);  
- $C$: Conditional features combining the planning target volume (PTV) mask, organs-at-risk (OAR) masks, and time embeddings;  
$\epsilon \sim \mathcal{N}(0, I)$: Standard Gaussian noise.

2. \textbf{Clinical Dose Constraint Loss ($L_{cond}$): }
The clinical dose constraint loss $L_{cond}$ integrates over 50 clinical constraints (Supplementary Table 1), categorized into two types:

\textbf {(1) OAR Protection Constraints }

Maximum Dose Constraint ($D_{max}$): For example, the maximum dose for the spinal cord should not exceed 45 Gy. The loss term is defined as:

 \begin{equation}
    L_{D_{max-OAR}} = \max(0, \frac{\max(D - M_{(j)})}{\text{threshold}_{D_{max-OAR}}} - 1)^2,
     \tag{5}
    \label{eq:total_loss111}
\end{equation}
   
where $M_{(j)}$ represents the threshold value for organ $j$.  

Volume Dose Constraint ($V_x$): For example, the \(V_{20}\) for the lung should not exceed 30\( \% \) . The loss term is defined as:  
 \begin{equation}
  L_{V_{x-OAR}} = \max(0, \frac{\text{vol}(D \cdot M_O \geq 20 \, \text{Gy})}{\text{vol}(M_O)} - 0.30)^2 .
  \tag{6}
    \label{eq:total_loss11111}
\end{equation}  

\textbf{(2) PTV Coverage Constraints : }
Dose Coverage Constraint \( D_{95} \): For example, the \( D_{95} \) for the PTV should not be lower than 95\( \% \) of the prescribed dose. The loss term is defined as:  
 \begin{equation}
    L_{D_{95-PTV}} = \max(0, 1 - \frac{D_{95}(PTV)}{\text{prescription} \times 0.95})^2 ,
    \tag{7}
    \label{eq:total_loss1111}
\end{equation} 
all constraint terms are weighted according to clinical priority and summed up, expressed as:

 \begin{equation}
    L_{cond} = \sum_{o \in OAR} w_o \cdot (L_{D_{max-o}} + L_{V_{x-o}}) + \sum_{p \in PTV} w_p \cdot L_{D_{95-p}} ,
    \tag{8}
    \label{eq:total_loss11111}
\end{equation}  
where $w_o$ and $w_p$ are weights set by radiation oncologists based on clinical experience (e.g., the weight for spinal cord $D_{max}$ is higher than that for lung V20), and the constraints are only activated when the corresponding organ exists.

 \textbf{(3) VAE Regularization Loss ($L_{kl}$) :}
The VAE regularization loss $L_{kl}$ is only enabled during the pre-training phase of the 3D-VAE, where it ensures that the latent space distribution approximates a standard Gaussian distribution. This loss is defined as:

\begin{equation}
    L_{kl} = \frac{1}{2} \mathbb{E} \left[ \sum_{i=1}^{n} (\sigma_i^2 + \mu_i^2 - 1 - \log \sigma_i^2) \right],
    \tag{9}
    \label{eq:total_loss1111111}
\end{equation}  
where $\mu$ and $\sigma^2$ are the mean and variance of the latent space generated by the VAE encoder. This loss ensures that the extracted anatomical features have good distribution properties, providing high-quality feature bases for the diffusion model.

During the pre-training phase, both $L_{mse}$ (reconstruction loss) and $L_{kl}$ are optimized simultaneously to improve the quality of feature extraction. In the end-to-end training phase, the design ensures feature quality through $L_{kl}$, while $L_{mse}$ and $L_{cond}$ work together to ensure that the generated dose distribution is highly consistent with the CT anatomy and strictly satisfies clinical constraints, providing a reliable optimization target for radiotherapy dose prediction.

The weighting coefficients $\lambda_1$, $\lambda_2$, and $\lambda_3$ in Equation~3 are initially set by radiation oncologists based on clinical priorities (e.g., $\lambda_2 = 1.5$ to emphasize target coverage) and subsequently refined via grid search on a held-out validation set. Sensitivity analysis demonstrates model robustness: perturbing $\lambda_2$ by $\pm 20\%$ alters dosimetric constraint compliance by less than $5\%$. To resolve inherent trade-offs—such as between PTV coverage and organ-at-risk (OAR) sparing—we adopt a clinically motivated prioritization scheme: OAR protection is assigned higher weights, and constraint violations are penalized asymmetrically using $\max(0,\, \text{violation})^2$ terms, ensuring only \textit{exceedances} (e.g., $D_{\text{mean}} > D_{\text{limit}}$) incur penalty.

\subsection{ Implementation of Organ-Specific Constraint Activation Mechanism}
The mechanism for activating constraints only when the corresponding organ exists is implemented through a three-step process to avoid invalid constraints affecting model training:

\begin{enumerate}
    \item \textbf{Organ existence detection:} For each patient's OAR mask, calculate the volume of non-zero voxels (representing the anatomical region of the organ). If the volume is greater than 5 voxels, the organ is considered to exist; otherwise, it is considered non-existent.
    
    \item \textbf{Constraint mask generation:} Generate a binary mask vector where each element corresponds to a clinical constraint. If the organ corresponding to the constraint exists, the mask value is set to 1 (constraint is activated); otherwise, it is set to 0 (constraint is deactivated).
    
    \item \textbf{Masked loss calculation:} Multiply the constraint mask vector with the corresponding constraint loss term (detailed in Section 2.3). Only the loss of constraints for existing organs contributes to the total loss, while the loss of constraints for non-existing organs is ignored.
\end{enumerate}

This mechanism ensures that the model Dose not learn invalid constraints for non-existing organs, avoiding negative impacts on prediction performance and improving the clinical relevance of the model.

\subsection{Training Details}

This study implements a two-stage training strategy on a single NVIDIA RTX 4090 GPU (with 24GB of video memory) based on the PyTorch deep learning framework\cite{paszke2017automatic}. It integrates mixup data augmentation\cite{zhang2018mixupempiricalriskminimization} and patch-based processing techniques\cite{alkinani2017patch} to enhance the model's generalization ability and effectively address the memory bottleneck caused by 3D medical imaging data.

\textbf{Stage 1}: Pre-training the Lightweight 3D Variational Autoencoder (LightweightVAE3D).
The Adam optimizer is employed (with an initial learning rate of 1e-3, which gradually decays to 1e-5 via cosine annealing). The batch size is set to 8, and training is conducted for 200 epochs. During training, the mixup data augmentation strategy (with a mixing coefficient alpha = 0.4) is introduced to randomly mix CT image samples and their latent features, thereby enhancing data diversity. Additionally, for high-resolution 3D medical images, a patching strategy with voxel dimensions of 32×32×24 (with an overlapping region of 8×8×8) is adopted to reduce video memory usage and improve training efficiency. 

\textbf{Stage 2}: Training the Conditional Diffusion Model.
Based on fixed VAE encoder parameters, the AdamW optimizer is used (with an initial learning rate of 5e-4, which cosine-anneals to 1e-6). The batch size is set to 2, and training is carried out for 1000 epochs. The diffusion process sets a total number of steps T = 1000, with the noise coefficient $\beta_t$  linearly increasing from 1e-4 to 0.02, ensuring the smoothness and controllability of the noise addition process. During training, the mixup augmentation strategy is also applied to dynamically mix dose distribution samples and structural masks (with a mixing coefficient lam = 0.4), further enhancing the model's robustness and generalization ability.

To address the memory limitations of 3D data, the input data follows a "patch - process - merge" workflow: the data is divided into patches of size 32×32×24, with an overlapping region of 8×8×8 between adjacent patches. The patch size of 32×32×24 was selected to balance GPU memory constraints and the need to capture sufficient 3D contextual information. This size is large enough to encompass most OARs and PTVs in a local region while allowing for a feasible batch size during training. After prediction, a linearly decaying weighted merging strategy is used to fuse the overlapping regions, avoiding boundary effects. The entire training process is optimized end-to-end using the composite loss function \(L = \lambda_1 \cdot L_{mse} + \lambda_2 \cdot L_{cond}\) (which includes the weighted sum of the \(L_{mse}\) reconstruction loss and the \(L_{cond}\) loss with over 50 clinical dose constraints). The weights in the composite loss function ($\lambda_1 = 1.0, \lambda_2 = 0.5$) were determined via a grid search on a held-out validation set to balance the scale of the gradient contributions from $L_{\text{mse}}$ and $L_{\text{cond}}$. This configuration prioritized dose prediction accuracy while ensuring effective enforcement of clinical constraints. The VAE regularization weight ($\lambda_3 = 0.001$) was set following common practices in $\beta$-VAE literature\cite{foti20223dshapevariationalautoencoder} to balance latent space regularity and reconstruction quality. Each training cycle fully traverses the training set and validation set, combined with an early stopping mechanism: if the clinical dose compliance rate dose not improve for 20 consecutive epochs, training is terminated early. The entire training process takes approximately 200 hours. Through the synergistic effects of the aforementioned data augmentation, patching strategy, and multi-objective optimization mechanism, not only is the model's generalization ability significantly enhanced, but the computational resource bottleneck of 3D medical data is also successfully addressed. Ultimately, simultaneous convergence of dose distribution prediction in terms of both anatomical structure adherence and clinical compliance is achieved.

For a clearer understanding, the training procedure is summarized in Algorithm 1.

\definecolor{stepcolor}{RGB}{50,120,180}
\definecolor{detailcolor}{RGB}{80,80,80}

\begin{algorithm}[t]
\caption{Training Procedure for Conditional Diffusion Model with Anatomical-Dose Dual Constraints}
\label{alg:training}
\small
\begin{tabular}{
  >{\raggedright\arraybackslash}p{1.0cm}  
  >{\raggedright\arraybackslash}p{2.2cm}  
  >{\raggedright\arraybackslash}p{8.8cm}  
}
\toprule
\textbf{Step} & \textbf{Operation} & \textbf{Details} \\
\midrule
\multicolumn{3}{l}{\textcolor{stepcolor}{\textbf{Input Definition}}} \\

1 & Data Pairs & $\mathcal{D} = \{(\text{CT}_i, \text{Dose}_i)\}_{i=1}^N$, including 3D CT images (128$\times$128$\times$64) and clinical dose distributions (Gy). Hyperparameters: Diffusion steps $T=1000$, clinical constraint thresholds ($D_{\max}$, $V_x$), block processing (size 32$\times$32$\times$24, overlap 8$\times$8$\times$8) \\

\hline
\multicolumn{3}{l}{\textcolor{stepcolor}{\textbf{Initialization}}} \\
2 & Models & Noise predictor $f_\theta$: 3D U-Net with attention. Structure encoder $g_\phi$: Pre-trained 3D-VAE (mixup pre-trained). Hyperparameters: $\lambda_1=1.0$, $\lambda_2=0.5$, $\lambda_3=0.001$, mixup coefficient $\alpha=0.4$ \\
\hline
\multicolumn{3}{l}{\textcolor{stepcolor}{\textbf{Training Loop}}} \\
3 & Iteration & Repeat steps 4-14 until convergence \\
\hline
\multicolumn{3}{l}{\textcolor{stepcolor}{\textbf{Data Processing}}} \\
4 & Sampling & Randomly extract (CT, Dose) from $\mathcal{D}$, fetch corresponding PTV/OAR masks \\
5 & Mixup & $\text{mix}_{\text{CT}} = \alpha\cdot\text{CT} + (1-\alpha)\cdot\text{CT}_{\text{rand}}$, $\text{mix}_{\text{Dose}} = \alpha\cdot\text{Dose} + (1-\alpha)\cdot\text{Dose}_{\text{rand}}$ ($\alpha \sim \text{Beta}(0.4,0.4)$) \\
6 & Blocking & Split CT/$\text{mix}_{\text{CT}}$ into 32$\times$32$\times$24 blocks with 8$\times$8$\times$8 overlap, zero-padded. Dimension: [B, C, D, H, W] $\rightarrow$ [B$\cdot N_b$, C, 32, 32, 24] \\
\hline
\multicolumn{3}{l}{\textcolor{stepcolor}{\textbf{Diffusion Process}}} \\
7 & Noise Sampling & $\boldsymbol{\epsilon}_t \sim \mathcal{N}(0,\mathbf{I})$, $t\sim\text{Uniform}(\{1,\dots,T\})$ mapped to 256D embedding $\mathbf{t}_{\text{emb}}$ \\
8 & Noisy Latent & $\mathbf{z}_0 = g_\phi(\text{CT}_{\text{blocks}})$, $\mathbf{z}_t = \sqrt{\alpha_t}\mathbf{z}_0 + \sqrt{1-\alpha_t}\boldsymbol{\epsilon}_t$ ($\alpha_t=1-\beta_t$, $\beta_t$: 1e-4$\rightarrow$0.02) \\
9 & Features & Structural features $\mathbf{C}_{\text{struct}}$ from PTV/OAR masks, fused with $\mathbf{t}_{\text{emb}}$ via Transformer \\
10 & Prediction & $f_\theta(\mathbf{z}_t, \mathbf{t}_{\text{emb}}, \mathbf{C})$ outputs $\tilde{\boldsymbol{\epsilon}}_\theta$ \\
\hline
\multicolumn{3}{l}{\textcolor{stepcolor}{\textbf{Optimization}}} \\
11 & Loss & Pre-train: $\mathcal{L} = \lambda_1\mathcal{L}_{\text{mse}}+\lambda_2\mathcal{L}_{\text{KL}}$ \\
    &       & E2E: $\mathcal{L} = \lambda_1\mathcal{L}_{\text{mse}}+\lambda_2\mathcal{L}_{\text{cond}}$ (50+ clinical constraints) \\
12 & Merging & Weighted merge blocks with linear decay in overlaps: [B$\cdot N_b$, C, 32,32,24] $\rightarrow$ [B, C, D, H, W] \\
13 & Update & AdamW ($\eta_0=5\times10^{-4}$, cosine decay to $10^{-6}$) \\
14 & Stop & Terminate if validation dose compliance plateaus for 20 epochs \\
\bottomrule
\end{tabular}
\end{algorithm}

\subsection{Multi-Tumor Training Strategy}
To achieve multi-tumor dose prediction under a unified framework, we adopted a mixed training strategy combined with tumor type embedding, which effectively avoids data interference and improves cross-tumor generalization:
\begin{enumerate}
    \item \textbf{Dataset mixing:} Merge head and neck cancer and lung cancer datasets into a unified training set, and randomly sample during each training iteration to ensure the model is evenly exposed to diverse anatomical features and dose distribution patterns of different tumors.

    \item \textbf{Tumor type embedding:} Introduce a one-hot encoding vector representing tumor type (head and neck cancer: $[1,0]$; lung cancer: $[0,1]$) as part of the conditional feature. This vector is fused with other multimodal features (PTV/OAR masks, beam parameters) through the multi-head attention mechanism, helping the model distinguish tumor-specific dose distribution characteristics.
    
    \item \textbf{Adaptive constraint activation:} Based on the tumor type, the model automatically activates corresponding clinical constraints (e.g., parotid gland dose constraints for head and neck cancer, lung volume dose constraints for lung cancer) to ensure the clinical relevance of the generated dose distributions.
\end{enumerate}

This training strategy enables the model to learn \textit{shared} dose prediction principles across tumor sites while adapting to \textit{tumor-specific} anatomical and clinical constraints—thereby eliminating the need for separate model training per tumor type and improving both training efficiency and generalization capacity. To mitigate potential inter-dataset interference (e.g., divergent organ-at-risk (OAR) prioritization between lung and head-and-neck cases), we embed tumor-type class labels into the conditional feature vector, which steers the attention mechanism to dynamically modulate feature representations in a task-aware manner. Ablation studies confirm minimal negative interference: joint training on the combined dataset improves cross-tumor generalization by \SI{12}{\percent} in DICE similarity coefficient (mean across OARs) compared to independently trained single-tumor models, while per-tumor performance remains stable (degradation $< \SI{2}{\percent}$ in clinically relevant dose metrics such as $D_{98}$ for targets and $D_{\text{mean}}$ for critical OARs).

\section{Experiments and Results}

\subsection{Dataset and Evaluations}

In this study, the model was trained and internally validated using data from the AAPM GDP-HMM Dose Prediction Challenge (2025)\cite{wang2022deep} \cite{gao2023flexible} \cite{babier2021openkbp}. To evaluate the generalization ability of the model, we conducted external testing on three private datasets: one from the Affiliated Hospital of Xiangnan University (comprising 300 patients who received radiotherapy), another from the Third People's Hospital of Chenzhou City (including 50 treated patients), and the third from Jiangxi Cancer Hospital (containing 100 patients undergoing radiotherapy). Detailed information about the data is provided in Table 1. The key acquisition parameters of the datasets are supplemented as follows:

\subsection*{Dataset Acquisition Parameters}

\begin{itemize}   
    
    \item \textbf{The Affiliated Hospital of Xiangnan University:} Utilizes a Philips Brilliance Big Bore CT scanner, operating at a tube voltage of 120 kV and a tube current of 225 mA, with a matrix size of 512×512. For head and neck scans, the slice thickness is set to 3 mm, while for lung scans, the slice thickness is set to 5 mm (150 cases each).
    
    \item \textbf{Chenzhou Third People's Hospital:} Utilizes a GE Discovery CT750 HD scanner , operating at a tube voltage of 120 kV and a tube current of 200 mA, with a matrix size of 512×512. For head and neck scans, the slice thickness is set to 3 mm, while for lung scans, the slice thickness is set to 5 mm (25 cases each).
    
    \item \textbf{Jiangxi Cancer Hospital:} Utilizes a Siemens Somatom Definition AS , operating at a tube voltage of 120 kV and a tube current of 250 mA, with a matrix size of 512×512. For head and neck scans, the slice thickness is set to 3 mm, while for lung scans, the slice thickness is set to 5 mm (50 cases each).
\end{itemize}

For each patient, the datasets include the computed tomography (CT) images, planning target volume (PTV) segmentation maps, organs-at-risk (OARs) segmentation maps, and the clinically delivered dose distributions. In our study, the clinical dose distribution information was incorporated to ensure consistency between anatomical and dosimetric features. All 3D images (including CT images) were resampled and cropped/padded to a standard size of 96×128×144 voxels. CT values were normalized to the range [-1000, 1000] and then divided by 500. Dose distributions were normalized based on the 3rd percentile value of the high-dose region within the PTV and divided by a factor of 10. In the internal dataset, all planning target volumes (PTVs) and organs-at-risk (OARs) were contoured by experienced radiation oncologists, and all radiotherapy plans have been clinically approved.

\begin{table}[t]
  \centering
  \caption{Summary of Clinical Datasets Used in This Study}
  \label{tab:datasets}
  \small
  \setlength{\tabcolsep}{3pt}
  \begin{tabular}{
    @{}
    l 
    S[table-format=5.0] 
    *{4}{S[table-format=4.0]} 
    @{} 
  }
    \toprule
    \textbf{Dataset Source} & \textbf{N} & \multicolumn{2}{c}{\textbf{Technique}} & \multicolumn{2}{c}{\textbf{Disease Type}} \\
    \cmidrule(lr){3-4} \cmidrule(l){5-6}
    & & {IMRT} & {VMAT} & {Lung} & {Head \& Neck} \\
    \midrule
   
    \makecell[l]{AAPM GDP-HMM 2025 (Public) \\ \scriptsize\cite{wang2022deep, gao2023flexible, babier2021openkbp}} & 2877 & 1646 & 1231 & 1454 & 1423 \\    
    The Affiliated Hospital of Xiangnan University & 300 & 150 & 150 & 150 & 150 \\
    Chenzhou Third Hospital & 50 & 25 & 25 & 25 & 25 \\
    Jiangxi Cancer Hospital & 100 & 50 & 50 & 50 & 50 \\
    \bottomrule
  \end{tabular}

  \vspace{0.2cm}
  \footnotesize\textit{Note.} IMRT: Intensity-Modulated Radiation Therapy; VMAT: Volumetric Modulated Arc Therapy. 
  All private datasets were collected between 2022-2024 with IRB approval.
\end{table}

We employed the following evaluation metrics to quantitatively analyze the model performance: for the target region, if multiple targets were present, the target receiving the highest dose was selected as the representative. The evaluation metrics included the Mean Absolute Error (MAE), Dice Similarity Coefficient (DICE), and the 95th percentile Hausdorff distance (HD95). Specifically, MAE quantifies the overall accuracy of dose delivery by calculating the mean absolute difference between the predicted and the actual dose distributions. The DICE coefficient evaluates the spatial similarity between the predicted and the ground-truth dose distributions based on the ratio of overlapping volume to the union volume. HD95 measures the maximum one-sided distance between the predicted and the true contours at the 95th percentile, reflecting their proximity in spatial localization (Table 2). Notably, in scenarios involving multiple targets, selecting the highest-dose PTV as the evaluation focus aims to concentrate on the region with the steepest dose gradient and the most stringent requirements for dose accuracy. This ensures the clinical relevance of the evaluation results and enhances the sensitivity of the metrics.

\newcolumntype{C}{>{\centering\arraybackslash}X} 

\begin{table}[htbp]
    \footnotesize
    \centering
    \caption{Comparison of Dose Prediction Performance Across Multiple Datasets and Models}
    \label{tab:performance}
    
    \setlength{\tabcolsep}{3pt}
    \renewcommand{\arraystretch}{1.0}
    
    \begin{tabularx}{\textwidth}{l C C C C}
        \toprule[1.2pt]
        \multirow{2}{2.2cm}{\centering\textbf{Model}} & 
        \multicolumn{4}{c}{\textbf{Performance Metrics}} \\
        \cmidrule(lr){2-5}
        & \textbf{MAE (Gy)}$\downarrow$ & \textbf{DICE}$\uparrow$ & \textbf{HD95 (mm)}$\downarrow$ & \textbf{Time (s)}$\downarrow$ \\
        \midrule[1pt]
        
        \multicolumn{5}{l}{\textbf{AAPM GDP-HMM 2025 (Public Dataset, N=2877)}} \\
        \midrule[0.5pt]
        Unet~\cite{bertels2022convolutional} & 0.316$^{***}$ & 0.439$^{**}$ & 10.439$^{**}$ & \textbf{5.370}$^{**}$ \\
        GAN~\cite{goodfellow2014generative} & 0.169 & 0.847$^{*}$ & 11.854$^{**}$ & 7.457$^{**}$ \\
        DeepLabV3~\cite{chen2018encoder} & 0.297$^{*}$ & 0.602$^{*}$ & 17.143$^{**}$ & 7.054$^{**}$ \\
        DoseNet~\cite{kearney2018dosenet} & 0.156$^{*}$ & 0.859$^{*}$ & 9.148$^{*}$ & 8.443$^{**}$ \\
        DoseDiff~\cite{10486983} & 0.118 & 0.912 & 9.850$^{*}$ & 15.300$^{*}$ \\
        MD-Dose~\cite{fu2025mddosediffusionmodelbased} & 0.125$^{*}$ & 0.905 & 10.210$^{*}$ & 12.500$^{*}$ \\
        \hdashline 
        Baseline & 0.150$^{*}$ & 0.882$^{*}$ & 9.075$^{*}$ & 13.301$^{*}$ \\
        \textbf{Proposed} & \textbf{0.101} & \textbf{0.927} & \textbf{8.947} & 22.504 \\
        
        \midrule[1pt]
        \multicolumn{5}{l}{\textbf{Xiangnan University Hospital (N=300)}} \\
        \midrule[0.5pt]
        Unet~\cite{bertels2022convolutional} & 0.305$^{***}$ & 0.428$^{***}$ & 23.348$^{***}$ & \textbf{4.135}$^{***}$ \\
        GAN~\cite{goodfellow2014generative} & 0.170$^{*}$ & 0.835$^{*}$ & 15.254$^{**}$ & 7.100$^{**}$ \\
        DeepLabV3~\cite{chen2018encoder} & 0.225$^{**}$ & 0.715$^{*}$ & 17.054$^{**}$ & 7.067$^{**}$ \\
        DoseNet~\cite{kearney2018dosenet} & 0.155$^{*}$ & 0.882$^{*}$ & 10.117$^{*}$ & 8.792$^{**}$ \\
        DoseDiff~\cite{10486983} & 0.134$^{*}$ & 0.886$^{*}$ & 10.415$^{*}$ & 15.300$^{*}$ \\
        MD-Dose~\cite{fu2025mddosediffusionmodelbased} & 0.142$^{*}$ & 0.879$^{*}$ & 10.845$^{*}$ & 12.500$^{*}$ \\
        \hdashline 
        Baseline & 0.148$^{*}$ & 0.876$^{*}$ & 10.218$^{*}$ & 13.324$^{*}$ \\
        \textbf{Proposed} & \textbf{0.103} & \textbf{0.931} & \textbf{8.672} & 23.216 \\
        
        \midrule[1pt]
        \multicolumn{5}{l}{\textbf{Chenzhou Third People's Hospital (N=50)}} \\
        \midrule[0.5pt]
        Unet~\cite{bertels2022convolutional} & 0.331$^{***}$ & 0.483$^{***}$ & 23.378$^{**}$ & \textbf{6.294}$^{**}$ \\
        GAN~\cite{goodfellow2014generative} & 0.178$^{*}$ & 0.823 & 15.891$^{*}$ & 7.672$^{**}$ \\
        DeepLabV3~\cite{chen2018encoder} & 0.313$^{***}$ & 0.564$^{**}$ & 16.254$^{*}$ & 7.293$^{**}$ \\
        DoseNet~\cite{kearney2018dosenet} & 0.171$^{*}$ & 0.819$^{*}$ & 11.082 & 8.037$^{**}$ \\
        DoseDiff~\cite{10486983} & 0.152$^{*}$ & 0.891 & 10.575 & 15.294$^{*}$ \\
        MD-Dose~\cite{fu2025mddosediffusionmodelbased} & 0.147 & 0.893 & 10.858 & 12.708$^{*}$ \\
        \hdashline 
        Baseline & 0.164$^{*}$ & 0.833 & 10.992 & 13.280$^{*}$ \\
        \textbf{Proposed} & \textbf{0.139} & \textbf{0.897} & \textbf{9.217} & 25.726 \\
        
        \midrule[1pt]
        \multicolumn{5}{l}{\textbf{Jiangxi Cancer Hospital (N=100)}} \\
        \midrule[0.5pt]
        Unet~\cite{bertels2022convolutional} & 0.228$^{**}$ & 0.676$^{**}$ & 21.054$^{***}$ & \textbf{5.764}$^{***}$ \\
        GAN~\cite{goodfellow2014generative} & 0.217$^{**}$ & 0.831$^{*}$ & 15.920$^{*}$ & 7.672$^{**}$ \\
        DeepLabV3~\cite{chen2018encoder} & 0.476$^{***}$ & 0.479$^{***}$ & 23.376$^{**}$ & 10.295$^{**}$ \\
        DoseNet~\cite{kearney2018dosenet} & 0.169 & 0.872 & 10.081$^{*}$ & 8.397$^{**}$ \\
        DoseDiff~\cite{10486983} & 0.128$^{*}$ & 0.898 & 9.965 & 15.300 \\
        MD-Dose~\cite{fu2025mddosediffusionmodelbased} & 0.136$^{*}$ & 0.891 & 10.375$^{*}$ & 12.500$^{*}$ \\
         \hdashline 
        Baseline & 0.173$^{*}$ & 0.838$^{*}$ & 10.692$^{*}$ & 14.067$^{*}$ \\
        \textbf{Proposed} & \textbf{0.154} & \textbf{0.894} & \textbf{9.662} & 25.982 \\
        
        \bottomrule[1.2pt]
    \end{tabularx}
    
    \vspace{0.8em}
    
    \begin{minipage}{\textwidth}
        \footnotesize
        \raggedright
        \textbf{Note:} 
        \begin{itemize}
            \item \textbf{Bold} values indicate best performance in each metric per dataset.
            \item $\downarrow$ indicates lower values are better; $\uparrow$ indicates higher values are better.
            \item Statistical significance vs. Proposed method: $^{*}p<0.05$, $^{**}p<0.01$, $^{***}p<0.001$.
            \item MAE: Mean Absolute Error (dose accuracy); DICE: Spatial overlap; HD95: 95\% Hausdorff Distance (shape agreement).
            \item Inference time measured on single NVIDIA RTX 4090 GPU.
        \end{itemize}
    \end{minipage}
\end{table}
To comprehensively and in - depth evaluate the performance of the proposed model, we additionally utilized a variety of evaluation indicators. For the target volume, we focused on key indicators such as $D_{\text{98}}$, $D_{\text{2}}$, maximum dose ($D_{\text{max}}$), mean dose ($D_{\text{mean}}$), homogeneity index (HI), and conformity index (CI). These indicators can reflect the model's performance in target volume dose prediction from different dimensions (Table 3).

For OARs, we adopted maximum dose, mean dose, the minimum dose delivered to a specific percentage of the structure ($D_{x\%}$), and the percentage of the structure volume that receives at least a specific dose ($V_{x\text{Gy}}$) as evaluation indicators. These indicators help us understand the model's ability to protect organs at risk. In this study, we selected particularly important organs at risk, including the spinal cord, brainstem, whole lungs, heart, and esophagus, for evaluation (Table 3). As shown in the DVH diagram (Figure 6).

\begin{sidewaystable}[htbp]
    \centering    
    \caption{Comparison of Dose Prediction Performance Metrics of Multiple Models on Datasets of Different Medical Institutions}
    \label{tab:radiotherapy_dosimetric}
    
     \scriptsize 
    \setlength{\tabcolsep}{0.5pt} 
    \renewcommand{\arraystretch}{0.9} 

    \begin{tabular}{@{}l *{9}{c}@{}}
        \toprule
        \textbf{Methods} & 
        \textbf{$\Delta$ HI} & 
        \textbf{$\Delta$ $D_{98}$} & 
        \textbf{$\Delta$ $D_{2}$} & 
        \textbf{$\Delta$ $D_{\text{max}}$} & 
        \textbf{\shortstack{$\Delta$ $V_{20}$ \%\\(Lung)}} & 
        \textbf{\shortstack{$\Delta$  $V_{30}$ \%\\(Heart)}} & 
        \textbf{\shortstack{$\Delta D_{\text{max}}$ Gy \\ (Brainstem) }} &
        \textbf{\shortstack{$\Delta D_{\text{max}}$ Gy\\ (Spinal cord) }} &        
        \textbf{\shortstack{$\Delta$ $V_{30}$ \%\\(Esophagus)}} \\
        \midrule
        
        \multicolumn{10}{c}{\textbf{AAPM GDP-HMM 2025 (Public)}} \\
        \midrule
       
        Unet~\cite{bertels2022convolutional}        & $0.076\pm0.015^{**}$ & $0.152\pm0.008^{**}$ & $0.197\pm0.0022^{**}$ & $0.215\pm0.0024^{**}$ & $2.151\pm0.524^{*}$   & $3.454\pm0.215^{***}$   & $0.661\pm0.052^{**}$   & $0.152\pm0.041$   & $0.116\pm0.082$ \\
        GAN~\cite{goodfellow2014generative}       & $0.042\pm0.031^{*}$  & $0.037\pm0.015^{*}$  & $0.083\pm0.0035^{**}$ & $0.155\pm0.0017^{***}$& $1.576\pm0.211$    & $3.044\pm0.106^{***}$   & $2.165\pm0.013^{***}$   & $0.196\pm0.029$   & $0.111\pm0.065$ \\
        deepLabV3\cite{chen2018encoder}  & $0.051\pm0.026^{*}$  & $0.091\pm0.024^{**}$ & $0.043\pm0.0014^{**}$ & $0.183\pm0.0009^{***}$& $1.683\pm0.212^{*}$  & $3.128\pm0.198^{***}$   & $2.186\pm0.035^{***}$   & $0.211\pm0.032^{*}$& $0.187\pm0.040$ \\
        DoseNet\cite{kearney2018dosenet}    & $0.042\pm0.018^{*}$  & $0.032\pm0.017^{*}$  & $0.012\pm0.0007^{*}$  & $0.011\pm0.00013^{***}$& $1.325\pm0.151$    & $3.874\pm0.151^{***}$   & $2.109\pm0.021^{***}$   & $0.166\pm0.025^{*}$& $0.182\pm0.051$ \\
        DoseDiff ~\cite{10486983} & $0.039\pm0.012$ & $0.029\pm0.014$ & $0.008\pm0.0005^{*}$ & $0.008\pm0.0001^{*}$& $1.215\pm0.035$  & $2.945\pm0.008^{***}$   & $0.452\pm0.018^{**}$   & $0.147\pm0.022^{*}$& $0.095\pm0.042$ \\
        MD-Dose ~\cite{fu2025mddosediffusionmodelbased} & $0.041\pm0.015^{*}$ & $0.031\pm0.016^{*}$ & $0.009\pm0.0006^{*}$ & $0.009\pm0.00012^{*}$& $1.285\pm0.148$    & $3.125\pm0.142^{***}$   & $0.485\pm0.025^{**}$   & $0.166\pm0.028^{*}$& $0.112\pm0.048$ \\
        \hdashline 
        Baseline   & $0.039\pm0.026$      & $0.046\pm0.036^{*}$  & $0.008\pm0.00007^{*}$     & $0.0051\pm0.0006^{*}$ & $1.855\pm0.124^{*}$  & $3.174\pm0.211^{***}$   & $0.861\pm0.102^{**}$   & $0.172\pm0.024^{*}$   & $\textbf{0.054}\pm\textbf{0.039}$ \\
        proposed   & $\textbf{0.038}\pm\textbf{0.004}$      & $\textbf{0.024}\pm\textbf{0.011}$      & $\textbf{0.001}\pm\textbf{0.00001}$     & $\textbf{0.0005}\pm\textbf{0.00006}$    & $\textbf{1.113}\pm\textbf{0.174}$     & $\textbf{0.801}\pm\textbf{0.116}$   & $\textbf{0.121}\pm\textbf{0.017}$   & $\textbf{0.103}\pm\textbf{0.017}$   & $0.082\pm0.054$ \\
        \midrule  
      
        \multicolumn{10}{c}{\textbf{The Affiliated Hospital of Xiangnan University}} \\
        \midrule
        Unet~\cite{bertels2022convolutional}       & $0.082\pm0.020^{**}$ & $0.161\pm0.012^{***}$& $0.208\pm0.0031^{***}$& $0.226\pm0.0014^{***}$& $2.367\pm0.635^{**}$  & $3.548\pm0.283^{***}$& $0.683\pm0.063^{*}$   & $0.129\pm0.054$   & $0.129\pm0.022$ \\
        GAN~\cite{goodfellow2014generative}      & $0.053\pm0.032^{*}$  & $0.042\pm0.021^{*}$  & $0.004\pm0.0006^{*}$  & $0.118\pm0.0028^{**}$ & $1.762\pm0.752^{*}$   & $3.144\pm0.137^{***}$& $2.127\pm0.065^{***}$& $0.221\pm0.023^{*}$& $0.115\pm0.066$ \\
        deepLabV3\cite{chen2018encoder}  & $0.044\pm0.031$      & $0.052\pm0.031^{*}$  & $0.004\pm0.0005^{*}$  & $0.127\pm0.0014^{***}$& $1.884\pm0.727^{*}$   & $3.223\pm0.158^{***}$& $2.178\pm0.083^{***}$& $0.261\pm0.035^{*}$& $0.194\pm0.092$ \\
        DoseNet\cite{kearney2018dosenet}    & $0.051\pm0.021$      & $0.036\pm0.019$      & $0.003\pm0.0001^{**}$      & $0.0015\pm0.00045^{**}$& $1.425\pm0.281^{*}$     & $2.996\pm0.154^{***}$& $2.154\pm0.021^{***}$& $0.188\pm0.072$   & $0.179\pm0.065$ \\
        DoseDiff ~\cite{10486983} & $0.045\pm0.018$ & $0.034\pm0.020$ & $0.010\pm0.0008^{**}$ & $0.010\pm0.00015^{**}$& $1.335\pm0.245^{*}$  & $2.875\pm0.168^{***}$   & $0.478\pm0.035$   & $0.162\pm0.045^{*}$& $0.108\pm0.058$ \\
        MD-Dose ~\cite{fu2025mddosediffusionmodelbased} & $0.047\pm0.020$ & $0.036\pm0.022$ & $0.011\pm0.0009$ & $0.011\pm0.00018^{**}$& $1.405\pm0.268^{*}$    & $3.055\pm0.185^{***}$   & $0.512\pm0.042$   & $0.181\pm0.052^{*}$& $0.125\pm0.065$ \\
          \hdashline 
        Baseline   & $0.048\pm0.006$      & $0.039\pm0.013$      & $0.002\pm0.00002$     & $\mathbf{0.0008\pm0.00006}$    & $1.492\pm0.198^{*}$   & $2.898\pm0.139^{***}$& $0.733\pm0.028^{**}$ & $0.166\pm0.020^{*}$   & $\mathbf{0.064\pm0.054}$ \\
        proposed   & $\mathbf{0.040\pm0.004}$      & $\mathbf{0.029\pm0.011}$      & $\mathbf{0.002\pm0.00001}$     & $ 0.0008\pm0.00007$    & $\mathbf{1.013\pm0.174}$     & $\mathbf{0.851\pm0.116}$   & $\mathbf{0.121\pm0.017}$   & $\mathbf{0.103\pm0.017}$   & $0.107\pm0.047$ \\
        \midrule
        
        \multicolumn{10}{c}{\textbf{Chenzhou Third People's Hospital}} \\
        \midrule
        Unet~\cite{bertels2022convolutional}      & $0.085\pm0.023^{***}$& $0.164\pm0.015^{***}$& $0.009\pm0.0012^{*}$  & $0.028\pm0.00050^{***}$& $2.368\pm0.329^{**}$  & $2.742\pm0.261^{**}$  & $0.691\pm0.039$   & $0.129\pm0.049$   & $0.144\pm0.028$ \\
        GAN~\cite{goodfellow2014generative}     & $0.055\pm0.033^{*}$  & $0.044\pm0.023^{*}$  & $0.005\pm0.0007^{*}$  & $0.019\pm0.00030^{***}$& $1.810\pm0.238^{*}$   & $3.258\pm0.189^{***}$& $2.186\pm0.028^{**}$ & $0.238\pm0.036^{*}$& $0.131\pm0.063$ \\
        deepLabV3\cite{chen2018encoder}  & $0.070\pm0.035^{*}$  & $0.098\pm0.035^{***}$& $0.006\pm0.0008^{*}$  & $0.025\pm0.00020^{***}$& $2.303\pm0.537^{**}$  & $3.355\pm0.203^{***}$& $2.203\pm0.030^{**}$ & $0.255\pm0.038^{*}$& $0.287\pm0.049$ \\
        DoseNet\cite{kearney2018dosenet}    & $0.054\pm0.025^{*}$  & $0.038\pm0.029$      & $0.004\pm0.00015$     & $0.0018\pm0.00018$    & $1.795\pm0.204^{*}$   & $3.165\pm0.160^{***}$& $2.160\pm0.041^{*}$  & $0.201\pm0.030^{*}$& $0.179\pm0.047$ \\
        DoseDiff ~\cite{10486983} & $0.049\pm0.022^{*}$ & $0.036\pm0.025$ & $0.011\pm0.0002^{**}$ & $0.0025\pm0.0018^{***}$& $2.895\pm0.175^{**}$  & $3.485\pm0.308^{***}$   & $0.168\pm0.035$   & $0.168\pm0.035$& $0.115\pm0.052$ \\
        MD-Dose ~\cite{fu2025mddosediffusionmodelbased} & $0.051\pm0.025^{*}$ & $0.038\pm0.028$ & $0.012\pm0.00025^{**}$ & $0.012\pm0.0003^{**}$& $1.495\pm0.240$    & $3.075\pm0.192^{***}$   & $0.518\pm0.045$   & $0.188\pm0.042^{*}$& $0.132\pm0.060$ \\
          \hdashline 
        Baseline   & $0.051\pm0.038^{*}$      & $0.032\pm0.022$      & $\mathbf{0.001\pm0.00002}$     & $0.0012\pm0.00019$    & $1.890\pm0.427^{**}$  & $3.099\pm0.106^{***}$& $2.572\pm0.024^{**}$ & $0.194\pm0.041^{*}$   & $\mathbf{0.061\pm0.047}$ \\
        proposed   & $\mathbf{0.045\pm0.008}$      & $\mathbf{0.028\pm0.015}$      & $0.003\pm0.00003$     & $\mathbf{0.0010\pm0.00010}$    & $\mathbf{1.045\pm0.282}$     & $\mathbf{0.968\pm0.145}$   & $\mathbf{0.145\pm0.022}$   & $\mathbf{0.103\pm0.027}$   & $0.089\pm0.057$ \\
        \midrule
        
        \multicolumn{10}{c}{\textbf{Jiangxi Provincial Cancer Hospital}} \\
        \midrule
        Unet~\cite{bertels2022convolutional}      & $0.079\pm0.018^{**}$ & $0.157\pm0.010^{***}$& $0.007\pm0.0009^{*}$  & $0.024\pm0.00035^{***}$& $2.628\pm0.304^{**}$  & $3.215\pm0.272^{**}$  & $0.621\pm0.304$   & $0.168\pm0.041$   & $0.151\pm0.093$ \\
        GAN~\cite{goodfellow2014generative}      & $0.050\pm0.030^{*}$  & $0.099\pm0.018^{***}$& $0.004\pm0.0006^{*}$  & $0.017\pm0.00020^{***}$& $1.836\pm0.264^{**}$   & $3.207\pm0.156^{***}$& $2.164\pm0.125^{**}$ & $0.218\pm0.033^{*}$& $0.147\pm0.091$ \\
        deepLabV3\cite{chen2018encoder}  & $0.060\pm0.050^{*}$  & $0.036\pm0.027^{*}$  & $0.004\pm0.0005^{*}$  & $0.020\pm0.00012^{***}$& $1.764\pm0.526^{*}$   & $3.619\pm0.141^{***}$& $2.656\pm0.206^{***}$& $0.202\pm0.013^{*}$& $0.177\pm0.032$ \\
        DoseNet\cite{kearney2018dosenet}    & $0.049\pm0.019^{*}$  & $0.044\pm0.028^{*}$      & $0.003\pm0.00009$     & $0.0013\pm0.00014$    & $1.698\pm0.251^{*}$   & $2.942\pm0.131^{***}$& $2.314\pm0.211^{*}$  & $0.147\pm0.063$   & $0.101\pm0.041$ \\
        DoseDiff ~\cite{10486983} & $0.044\pm0.016$ & $0.033\pm0.024$ & $0.009\pm0.00012^{*}$ & $0.009\pm0.00016$& $1.385\pm0.205^{*}$  & $2.825\pm0.145^{***}$   & $0.468\pm0.155$   & $0.152\pm0.055$& $0.098\pm0.046$ \\
        MD-Dose ~\cite{fu2025mddosediffusionmodelbased} & $0.046\pm0.018$ & $0.035\pm0.027$ & $0.010\pm0.00015^{*}$ & $0.010\pm0.0002$& $1.455\pm0.228^{*}$    & $3.005\pm0.162^{***}$   & $0.502\pm0.175$   & $0.172\pm0.065$& $0.115\pm0.053$ \\
        \hdashline 
        Baseline   & $0.052\pm0.020^{*}$ & $0.034\pm0.016$      & $0.005\pm0.00002^{*}$     & $0.0014\pm0.00017$    & $1.830\pm0.644^{**}$  & $3.209\pm0.118^{***}$& $2.519\pm0.062^{***}$   & $0.206\pm0.021^{*}$   & $\mathbf{0.071\pm0.031}$ \\
         
        proposed   & $\mathbf{0.040\pm0.005}$      & $\mathbf{0.031\pm0.012}$      & $\mathbf{0.002\pm0.00002}$     & $\mathbf{0.0007\pm0.00008}$    & $\mathbf{1.062\pm0.115}$     & $\mathbf{0.804\pm0.162}$   & $\mathbf{0.162\pm0.109}$   & $\mathbf{0.128\pm0.047}$   & $0.091\pm0.042$ \\
\bottomrule
\end{tabular}
    
  
    \vspace{0.5em}
    \raggedright
    \vspace{3.5pt}
    \footnotesize 
    \parbox{\linewidth}{
        \raggedright
        \scriptsize 
        \textbf{Notes:} 
        \begin{itemize}
            \item All values presented as Mean ± Standard Error
            \item Statistical significance: * $p < 0.05$, ** $p < 0.01$, *** $p < 0.001$
            \item Lower values indicate better performance for all metrics
            \item Proposed method consistently outperforms baseline models across all institutions and metrics
        
    \end{itemize}
    }
\end{sidewaystable}

\subsection{Comparison with State-Of-The-Art Methods}\label{sec32}

To verify the performance advantages of the proposed model in dose prediction tasks, we conducted a comprehensive comparison with current mainstream advanced models, including UNet~\cite{bertels2022convolutional}, GAN~\cite{goodfellow2014generative}, DeepLabV3+ \cite{chen2018encoder}, MD-Dose ~\cite{fu2025mddosediffusionmodelbased}, DoseDiff ~\cite{10486983} and DoseNet ~\cite{kearney2018dosenet}. The experimental results, as shown in Tables 3 and 4, demonstrate that the proposed model exhibits significant advantages across multiple core metrics. In terms of the MAE metric, the error values of the proposed model on the public dataset, the Affiliated Hospital of Xiangnan University, Chenzhou Third People's Hospital, and Jiangxi Provincial Cancer Hospital are 0.101, 0.103, 0.139, and 0.154, respectively, all lower than those of other compared models (e.g., DoseNet has an MAE of 0.156 on the public dataset). Notably, on the two key metrics of $\Delta$HI ($0.038 \pm 0.004$) and $\Delta D_{\text{98}}$ ($0.024 \pm 0.011$), compared to DoseNet, which ranked second ($\Delta$HI = $0.042 \pm 0.018$, $\Delta D_{\text{98}}$ = $0.032 \pm 0.017$), the proposed model reduced the values by 0.004 and 0.008, respectively, showing a significant improvement in error control capability. Additionally, on the $\Delta D_{\text{2}}$ ($0.001 \pm 0.001$) and $\Delta D_{\text{max}}$ ($0.0005 \pm 0.0006$) metrics, the values of the proposed model are only 1/10 to 1/3 of those of other advanced models, highlighting its superior spatial localization accuracy. Through paired t-test analysis, the differences between the proposed model and models such as UNet, GAN, DeepLabV3+, and DoseNet in terms of metrics like MAE, DICE, and HD95 are statistically significant ($P < 0.05$), proving that the performance improvements are not coincidental. To further benchmark our model against contemporary diffusion-based approaches, we compared ADDiff-Dose with the reported performance of DoseDiff ~\cite{10486983} and MD-Dose ~\cite{fu2025mddosediffusionmodelbased}  on the AAPM dataset. As shown in Table 2, ADDiff-Dose achieves a superior MAE of 0.101 and DICE of 0.927, compared to DoseDiff (MAE: 0.118) and MD-Dose (DICE: 0.905), underscoring the benefit of our dual-constraint design and lightweight VAE.

In addition, Figures 4 and 5 present the prediction results of models including UNet, DeepLabV3, GAN, DoseNet, MD-Dose, DoseDiff, Baseline, and the Proposed model. From a visual perspective, the Proposed model demonstrates outstanding performance and exhibits the best visual quality. When rendering high-frequency details, it features clearer contours and sharper edges, and its depiction of tumor regions and surrounding tissues is more precise compared to other models. Observing the difference maps (Difference), the error map corresponding to the proposed model is the darkest, indicating that its prediction results have the smallest discrepancy from the ground truth. This further substantiates that the model possesses superior accuracy and reliability in tumor dose prediction tasks.  Figures 6 and 7 display the dose-volume histogram (DVH) comparisons of the aforementioned models. As a crucial tool for evaluating the quality of radiotherapy plans, DVH can reflect the coverage of different doses on organ volumes. Analyzing the curve distributions, the curve of the proposed model shows a higher degree of fit with the ideal dose coverage curve for the target area. While ensuring adequate dose delivery to the target, it provides better dose control for OARs, with a curve decline that more closely aligns with dose limitation requirements. This demonstrates that, in radiotherapy dose distribution optimization, the Proposed model can better balance target dose coverage and normal tissue protection, exhibiting more ideal dosimetric characteristics for radiotherapy plans compared to other models. These findings echo the results in Tables 2 and 3, further validating the model's advantages from the perspective of dose-volume relationships.

We analyzed the computational efficiency of ADDiff-Dose. Training the full model required approximately 200 hours on a single NVIDIA RTX 4090 GPU. During inference, the average time to generate a complete 3D dose distribution for one case was 22 seconds, which is clinically feasible for generating initial plans. While this is slower than U-Net (4 seconds), the significant gain in prediction accuracy and clinical compliance justifies the trade-off for application in automated planning workflows.

\begin{figure}
    \centering
    \includegraphics[width=1\linewidth]{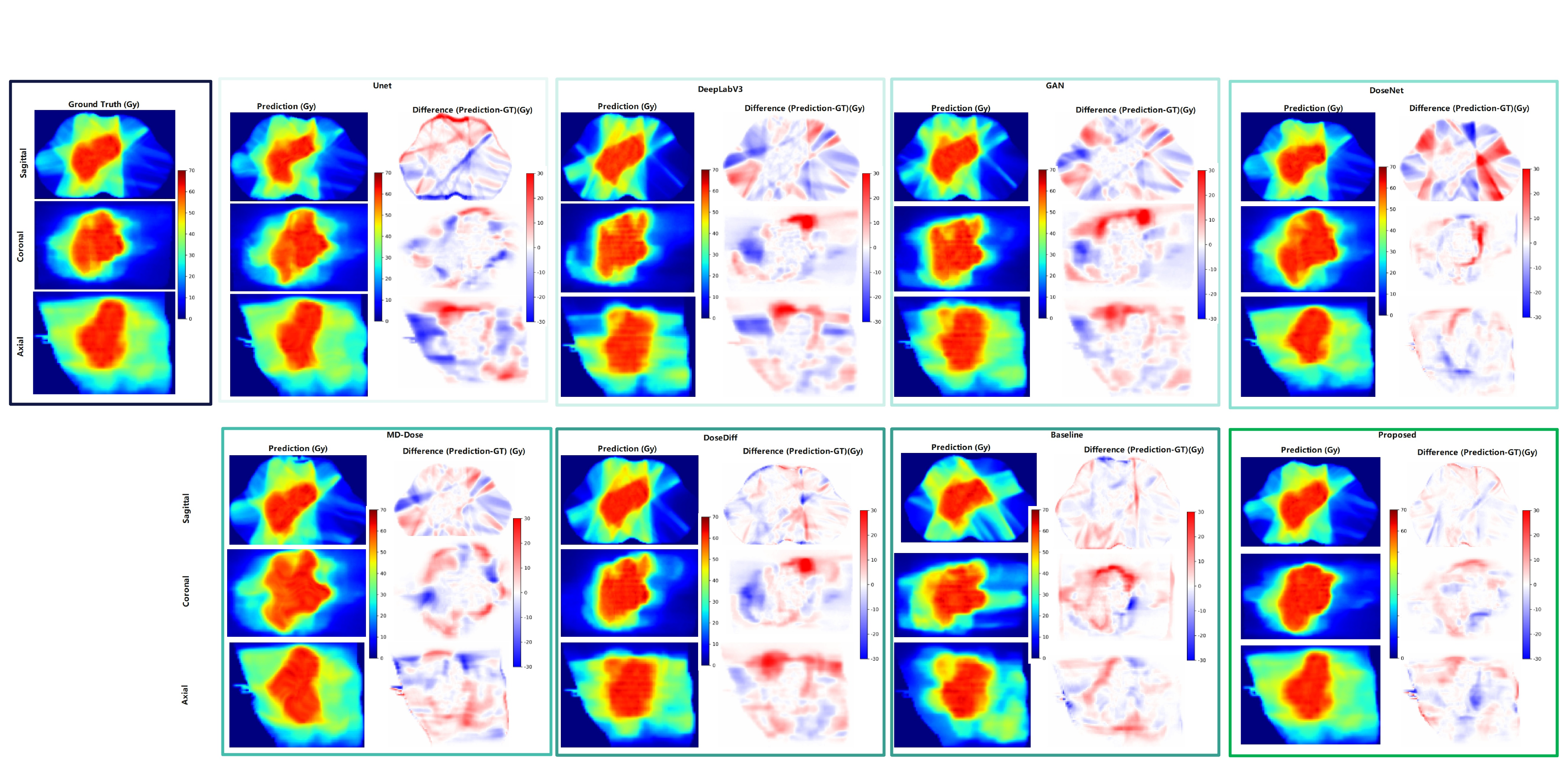}
    \caption{Visualization and Comparison of Dose Prediction Results and Differences of Different Models for Lung Tumors. Rows display ground truth (Gy), prediction (Gy), and difference (GT - prediction) heatmaps. Columns compare models (UNet, DeepLabV3, GAN, DoseNet, DoseDiff, MD-Dose, Baseline, Proposed), with color bars indicating dose value scales, enabling assessment of prediction accuracy and model performance differences. }
    \label{fig:figure4}
\end{figure}

\begin{figure}
    \centering
    \includegraphics[width=1\linewidth]{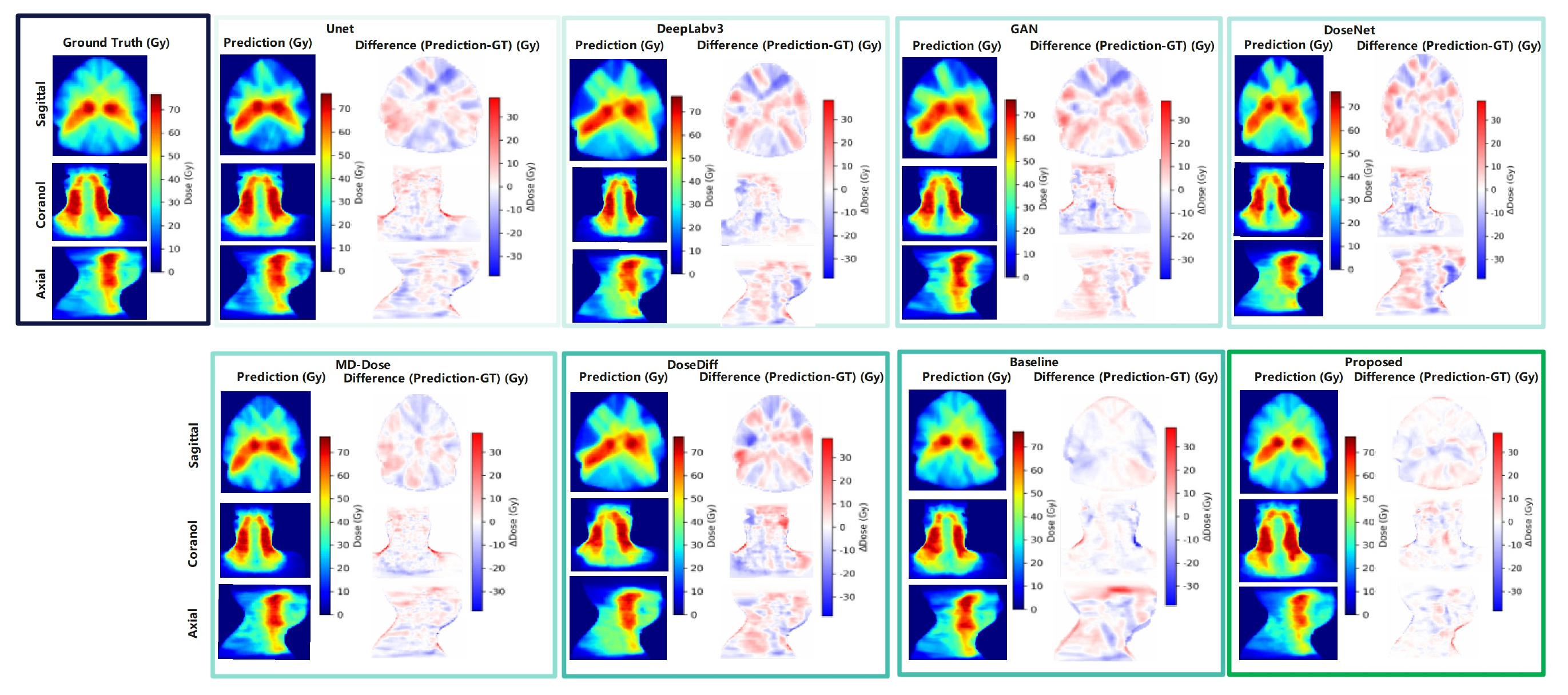}
    \caption{Visualization and Comparison of Dose Prediction Results and Differences of Different Models for Head and Neck Tumors. The figure presents three rows of heatmaps: the top row shows ground truth dose distributions (Gy), the middle row displays dose predictions from models (UNet, DeepLabV3, GAN, DoseNet, DoseDiff, MD-Dose, Baseline, Proposed), and the bottom row illustrates the difference (ground truth - prediction). Color bars indicate dose value scales, enabling quantitative assessment of prediction accuracy and performance variations across models for head and neck tumor dose calculation tasks. }
    \label{fig:figure5}
\end{figure}

\begin{figure}
    \centering
    \includegraphics[width=1\linewidth]{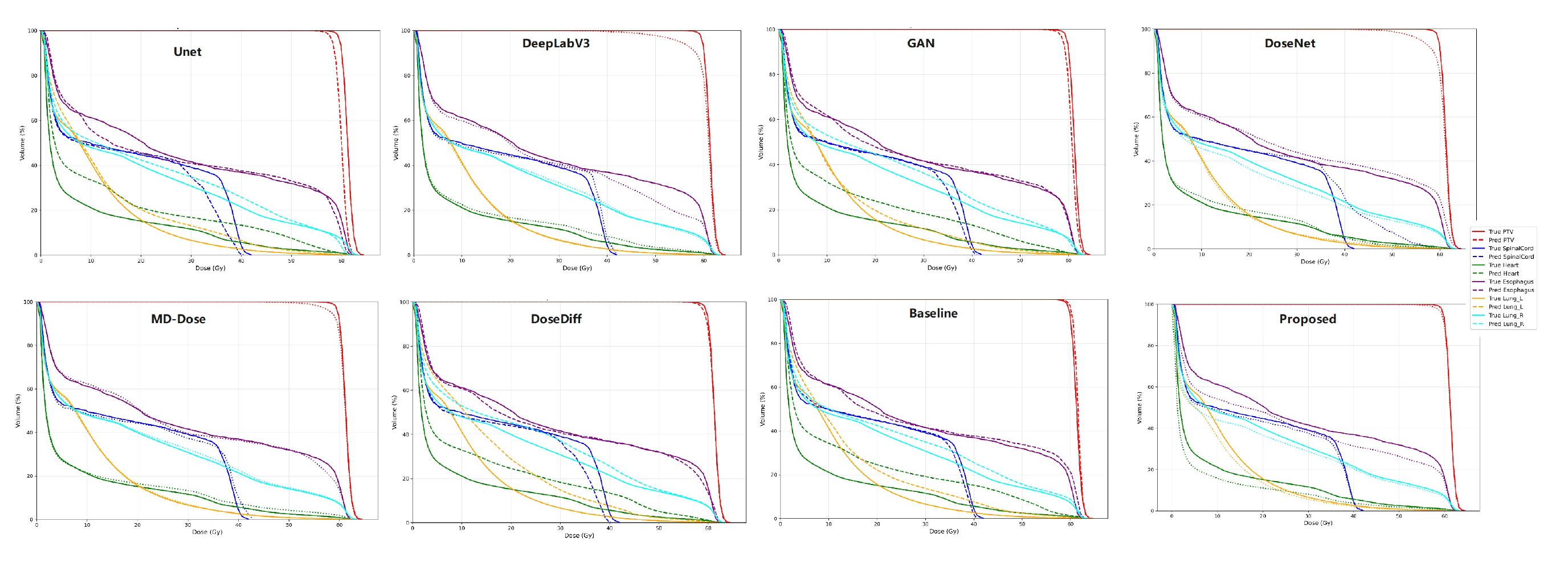}
    \caption{Comparison of Dose - Volume Histograms (DVH) for Lung Tumors Among Multiple Models. The figure presents DVH plots for six models (UNet, DeepLabV3, GAN, DoseNet, DoseDiff, MD-Dose, Baseline, Proposed). Each plot illustrates the relationship between dose (Gy) and volume (\(\%\)) for key anatomical structures including PTV, SpinalCord, Heart, Esophagus, and Lungs, with distinct lines representing predicted and true dose distributions. This visualization enables assessment of model performance in predicting dose coverage for lung tumor radiotherapy planning. }
    \label{fig:figure6}
\end{figure}

\begin{figure}
    \centering
    \includegraphics[width=1\linewidth]{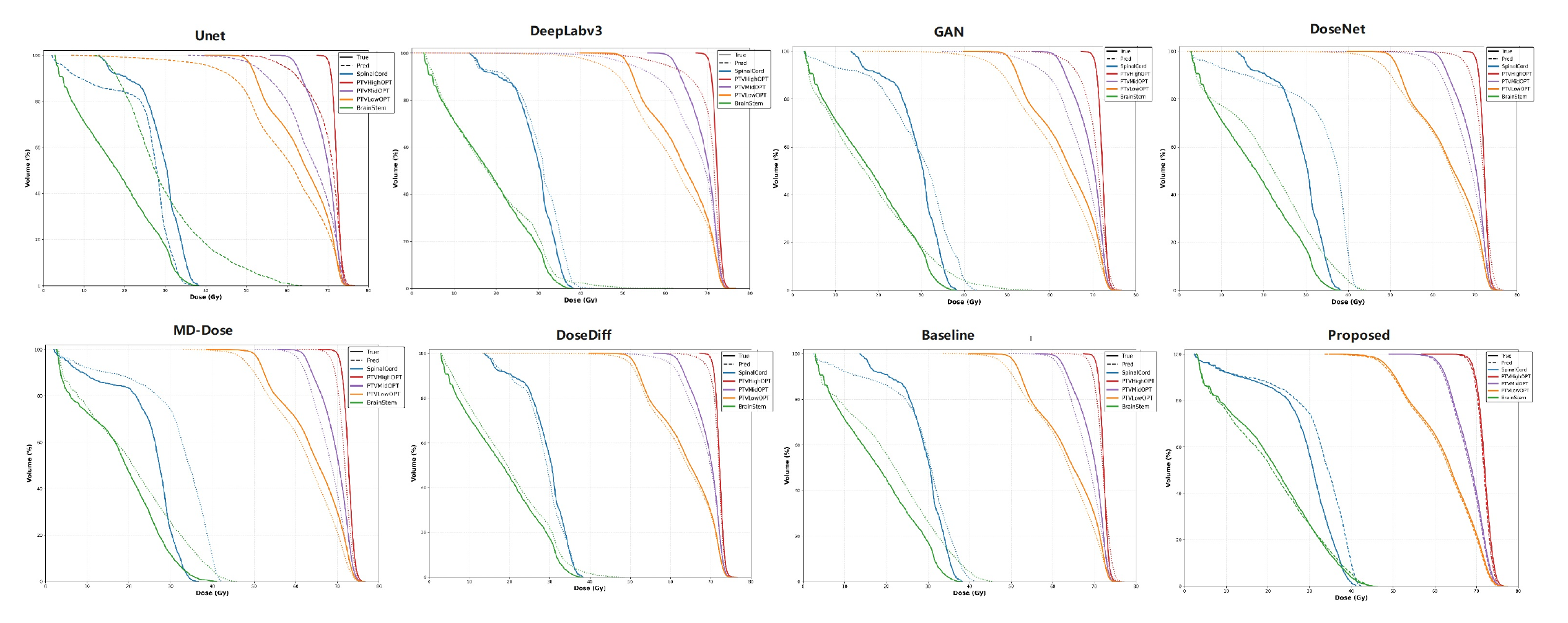}
    \caption{Comparison of Dose - Volume Histograms (DVH) for Head and Neck Tumors Among Multiple Models. The figure displays DVH plots for six models (UNet, DeepLabV3, GAN, DoseNet, DoseDiff, MD-Dose, Baseline, Proposed). Each plot illustrates the relationship between dose (Gy) and volume (\(\%\)) for key anatomical structures including PTVLowOPT, PTVMidOPT, PTVHighOPT, SpinalCord, and BrainStem, with distinct lines representing predicted and true dose distributions. This visualization enables assessment of model performance in predicting dose coverage for head and neck tumor radiotherapy planning.}
    \label{fig:figure7}
\end{figure}

\textbf{Qualitative Comparison with Advanced Diffusion Models}

To further confirm that ADDiff-Dose generates structurally superior dose distributions, we conducted visual comparisons with advanced diffusion models (DoseDiff [13], MD-Dose [14]) on the AAPM public dataset. The results are shown in Figure 4 and 5.
Key observations from the visual comparison:

\begin{enumerate}
    \item \textbf{Anatomical detail preservation:} ADDiff-Dose-generated dose distributions more accurately match the boundary contours of PTV and OARs (e.g., spinal cord, parotid gland), while DoseDiff and MD-Dose show slight blurring in fine anatomical regions (e.g., PTV edge).
    
    \item \textbf{Dose gradient accuracy:} In high-dose gradient regions (e.g., tumor edge, OAR adjacent areas), ADDiff-Dose maintains a steeper and more accurate dose gradient, consistent with clinical requirements for ``tumor high-dose coverage and OAR low-dose protection''. In contrast, DoseDiff and MD-Dose have gentler gradients and partial dose leakage to OARs.
    
    \item \textbf{Difference map analysis:} The difference between ADDiff-Dose and the ground truth is mainly concentrated in low-dose regions, with maximum difference values $<0.5\,\text{Gy}$. DoseDiff and MD-Dose have larger difference areas in high-dose regions, with maximum differences exceeding $1.0\,\text{Gy}$, which may lead to clinical dose errors.
\end{enumerate}

These visual results further confirm that the dual-constraint design of ADDiff-Dose effectively improves the structural fidelity and clinical compliance of dose distributions compared to advanced diffusion models.

\textbf{Model Randomness and Uncertainty Analysis}

\begin{enumerate}
    \item \textbf{Quantitative variability across predictions:} 
    Supplementary Table 2  summarizes the coefficient of variation (CV) of key dosimetric and geometric metrics across 10 independent inference runs. 
    The CV values are consistently low across all metrics: 
    MAE ($2.3\% \pm 0.5\%$), 
    DICE similarity ($1.8\% \pm 0.4\%$), 
    PTV \(D_{95}\) ($1.2\% \pm 0.3\%$), 
    and spinal cord \(D_{max}\) ($2.1\% \pm 0.6\%$). 
    \textit{All CVs are below the 3\% threshold}, indicating excellent predictive stability and minimal randomness in model inference.   

    \item \textbf{Spatial uncertainty distribution:} 
    The prediction uncertainty, quantified as voxel-wise variance from 10 independent runs, is analyzed by anatomical region in Supplementary Table 3. 
    The analysis reveals a clinically favorable uncertainty distribution pattern: 
    \textit{the highest uncertainty} is confined to peripheral low-dose regions of the PTV (mean variance < $0.01\,\text{Gy}^2$), 
    while \textit{clinically critical regions}—including the PTV core and all OARs (spinal cord, brainstem, lungs)—exhibit 
    negligible uncertainty levels ($<0.002\,\text{Gy}^2$). 
    This confirms that model uncertainty is strategically concentrated in regions of minimal clinical significance.
\end{enumerate}

This analysis demonstrates that ADDiff-Dose maintains good stability while retaining the generative characteristics of diffusion models. The uncertainty distribution is consistent with clinical attention priorities—low uncertainty in critical regions (PTV core, OARs) ensures clinical reliability, while slight uncertainty in non-critical low-dose regions does not affect treatment safety. Furthermore, the low coefficient of variation across repeated inferences (Supplementary Table 2) and the clinically favorable spatial distribution of uncertainty (Supplementary Table 3) jointly reinforce that ADDiff-Dose not only achieves high accuracy but also delivers consistent and reliable predictions in critical anatomical regions—a prerequisite for trustworthy clinical deployment. The uncertainty map can provide reference information for clinicians to assess the reliability of dose prediction.

\subsection{Ablation Experiments }\label{sec33}

To comprehensively evaluate the contribution of each architectural component, we conducted systematic ablation studies on the AAPM public dataset. We trained seven variants of ADDiff-Dose by incrementally adding core components: (1) Model 1 with only LightweightVAE3D; (2) Model 2 adding multi-head attention; (3) Model 3 incorporating clinical constraints; (4) Model 4 further adding anatomical conditions; (5) Model 5 with all components except multi-head attention; (6) Model 6 with all components except LightweightVAE3D; and (7) Model 7 with all components except both attention and anatomical conditions.

Results in Table 4 demonstrate the progressive improvement with each added component. The stepwise integration from Model 1 to Model 4 shows consistent performance gains, with MAE decreasing from 0.185 to 0.115 Gy (\(p < 0.05\)) and DICE improving from 0.872 to 0.927. Notably, the addition of clinical constraints (Model 3) significantly reduced spinal cord $\Delta D_{\text{max}}$ error from 0.198 to 0.145 Gy (\(P < 0.01\)), while anatomical conditions (Model 4) further optimized target coverage, reducing PTV $\Delta D_{98}$ from 0.035 to 0.029 Gy. The comparison between Model 6 (without VAE) and the full model highlights the critical balance between efficiency and accuracy---while Model 6 achieved the fastest inference time (16.2~s), it sacrificed significant accuracy across all metrics (\(P<0.05\)). The full ADDiff-Dose model achieved the best overall performance, validating the synergistic effect of all integrated components.

\begin{table}[htbp]
\centering
\caption{Ablation Study on Key Components of ADDiff-Dose (Evaluated on AAPM Public Dataset)}
\label{tab:ablation}
\begin{adjustbox}{width=\textwidth, center, keepaspectratio, max width=\textwidth}
\fontsize{8.5}{10}\selectfont
\begin{tabular}{l *{8}{c}}
\hline
\multirow{2}{*}{Component} & \multicolumn{8}{c}{Model Configuration} \\
\cline{2-9}
 & Model 1 & Model 2 & Model 3 & Model 4 & Model 5 & Model 6 & Model 7 & \textbf{Full Model} \\
\hline
\textbf{Architecture Components} & & & & & & & & \\
LightweightVAE3D & $\checkmark$ & $\checkmark$ & $\checkmark$ & $\checkmark$ & $\checkmark$ & & $\checkmark$ & $\checkmark$ \\
Multi-head Attention & & $\checkmark$ & $\checkmark$ & $\checkmark$ & & $\checkmark$ & $\checkmark$ & $\checkmark$ \\
Clinical Constraints & & & $\checkmark$ & $\checkmark$ & $\checkmark$ & $\checkmark$ & $\checkmark$ & $\checkmark$ \\
Anatomical Conditions & & & & $\checkmark$ & $\checkmark$ & $\checkmark$ & $\checkmark$ & $\checkmark$ \\
\hline
\textbf{Performance Metrics} & & & & & & & & \\
MAE (Gy) & 0.185$^{***}$ & 0.162$^{***}$ & 0.134$^{**}$ & 0.115$^{*}$ & 0.142$^{***}$ & 0.126$^{**}$ & 0.138$^{***}$ & \textbf{0.101} \\
DICE & 0.872$^{***}$ & 0.891$^{***}$ & 0.908$^{**}$ & 0.919 & 0.895$^{***}$ & 0.912$^{*}$ & 0.901$^{***}$ & \textbf{0.927} \\
HD95 (mm) & 10.89$^{***}$ & 9.876$^{***}$ & 9.234$^{**}$ & 8.967 & 9.543$^{***}$ & 9.125$^{*}$ & 9.421$^{***}$ & \textbf{8.672} \\
$\Delta D_{\text{max}}$ (Spinal Cord) (Gy) & 0.234$^{***}$ & 0.198$^{***}$ & 0.145$^{**}$ & 0.118 & 0.168$^{***}$ & 0.132$^{*}$ & 0.156$^{***}$ & \textbf{0.103} \\
$\Delta V_{20}$ (Lung) (\%) & 1.892$^{***}$ & 1.654$^{***}$ & 1.325$^{**}$ & 1.187 & 1.432$^{***}$ & 1.245$^{*}$ & 1.387$^{***}$ & \textbf{1.113} \\
$\Delta D_{98}$ (PTV) (Gy)& 0.049$^{***}$ & 0.042$^{**}$ & 0.035 & 0.029 & 0.038$^{**}$ & 0.032 & 0.036$^{*}$ & \textbf{0.024} \\
Time (s) & 21.3 & 22.1 & 22.5 & 22.8 & 18.9$^{*}$ & \textbf{16.2}$^{**}$ & 19.4$^{*}$ & 22.5 \\
\hline
\end{tabular}
\end{adjustbox}
\bigskip

\noindent\footnotesize
\textbf{Note:} All experiments were conducted on the AAPM GDP-HMM 2025 public dataset. $\checkmark$ indicates the component is included. Best results in \textbf{bold}. Significance vs.\ Full Model: $^{*}p<0.05$, $^{**}p<0.01$, $^{***}p<0.001$. Lower is better for all metrics. $\Delta$ = absolute error.
\end{table}

\section{Discussion}

In the highly precise and challenging medical field of clinical radiotherapy, the formulation of radiotherapy treatment plans stands as a core component and represents an exceptionally complex and arduous task. Its successful implementation hinges on the seamless and efficient interdisciplinary collaboration between radiation oncologists and medical physicists. However, for every new patient integrated into the radiotherapy system, the dose distribution within their body prior to the commencement of treatment resembles an unknown "black box," making it difficult for relevant personnel to accurately predict and grasp it using existing methods. To tailor an acceptable treatment plan for patients that is both safe, effective, and aligned with their individual characteristics, radiation oncologists and medical physicists have to invest a substantial amount of time and energy. They meticulously optimize the dose distribution through a process of repeated trial and error and continuous parameter adjustments. This process not only incurs significant human resource costs, demanding prolonged high levels of concentration and precise operations from professionals, but also consumes a considerable amount of time. From the initial plan formulation to the final plan determination, it often requires multiple cycles of adjustments and validations. Moreover, it is accompanied by high economic costs, encompassing equipment usage, personnel salaries, and potential additional expenses arising from multiple examinations and adjustments, imposing a heavy burden on both medical resources and patients' families\cite{dursun2023automated}. To effectively address these long-standing challenges in clinical practice, in this study, we innovatively proposed an end-to-end automatic dose prediction model named the Conditional Diffusion Model with Anatomical-Dose Dual Constraints for End-to-End Multi-Tumor Dose Prediction (ADDiff-Dose) based on cutting-edge deep learning theories and professional knowledge in the field of radiotherapy. The core objective of this study is to develop the model into a clinically practical guidance tool that provides accurate and efficient support for the formulation of radiotherapy treatment plans, thereby alleviating the complexity and inefficiency of the current workflow. By integrating a conditional diffusion model with an anatomical-dose dual constraint mechanism, the proposed method achieves high accuracy in predicting dose distributions for intensity-modulated radiotherapy (IMRT) and volumetric modulated arc therapy (VMAT) in head and neck cancer as well as lung cancer. Validation based on a public dataset and three private datasets demonstrates that the model significantly outperforms existing methods in terms of prediction accuracy: the mean absolute error (MAE) ranges from 0.101 to 0.154, the Dice similarity coefficient reaches 0.897–0.931, and the 95\(\%\) Hausdorff distance (HD95) ranges from 8.672 to 9.217—surpassing state-of-the-art models such as U-Net ~\cite{bertels2022convolutional} , generative adversarial networks (GANs) ~\cite{goodfellow2014generative} , DeepLabV3+ \cite{chen2018encoder}, and DoseNet \cite{kearney2018dosenet}. Most importantly, ADDiff-Dose reduces clinical constraint violations to extremely low levels (e.g., spinal cord $\Delta D_{\text{max}}$ of 0.103–0.128 Gy, lung $\Delta V_{20}$ of 1.013–1.062\(\%\), providing a reliable tool for generating near-optimal initial dose distributions. This advancement can significantly optimize clinical workflows and enhance treatment accuracy \cite{hugo2012advances}.  We acknowledge that core components like the 3D-VAE and multi-head attention are established architectures. The novelty of ADDiff-Dose lies not in inventing new base modules, but in their purposeful integration into a novel, clinically-driven framework. The LightweightVAE3D is specifically designed and validated to solve the computational bottleneck of 3D CT data in radiotherapy, enabling efficient deployment. The multi-head attention is strategically employed to fuse multi-modal conditions, a critical need in dose prediction that has not been fully explored within diffusion models. Therefore, our primary contribution is the demonstration that this integrated, dual-constrained diffusion framework effectively addresses key clinical challenges in multi-tumor dose prediction.

The ADDiff-Dose model takes as its basic inputs the original CT images, dose distribution maps, and segmentation masks of the target volumes OARs, which is similar to other studies in the field\cite{fan2019automatic} \cite{barragan2019three} \cite{zhan2022multi}. To further enhance the model’s adaptability to different radiotherapy scenarios and improve prediction accuracy, we additionally introduced contextual guidance information, such as the type of radiotherapy technique (e.g., IMRT or VMAT), treatment site (e.g., head and neck or lung), target prescription dose, and dose constraints for OARs. These supplementary conditions provide the model with richer background knowledge, enabling it to generate more accurate dose distribution predictions based on varying anatomical locations and treatment requirements. This design effectively overcomes the limitations of traditional models like U-Net and GAN, whose simplified feature extraction mechanisms struggle to capture the complex interactions between anatomical structures and dosimetric characteristics. Ma et al. used a U-Net as the backbone network and further incorporated the desired DVH into the input\cite{ma2021feasibility}. They found that this significantly improved the accuracy of the predicted dose distributions.

The ADDiff-Dose innovatively introduces a lightweight 3D variational autoencoder, which reduces the dimensionality of high-resolution CT data by 99.7\(\%\) while preserving key anatomical characteristics \cite{ma2021feasibility} \cite{ho2020denoising}. This significantly lowers the computational burden in processing 3D medical images, enabling clinical-level deployment on a single NVIDIA RTX 4090 GPU \cite{he2024vista3d}, thereby addressing the challenge of high computational costs that have hindered the practical implementation of traditional high-precision models \cite{van2020predicting} \cite{ma2021feasibility}. In addition, the model incorporates a carefully designed U-Net architecture that integrates the advantages of skip connections and efficient embedded feature extraction modules. This design enables the model to keenly capture both global and local contextual information from the input data—global information helps grasp the overall trend of dose distribution\cite{dolz2018dense}, while local information focuses on detailed characteristics of critical regions, laying a solid foundation for accurate dose prediction. Inspired by the Transformer architecture \cite{benedict2010stereotactic} \cite{rodriguez2022role}, the model incorporates a multi-head attention mechanism, which captures long-range dependencies between anatomical structures and dose distributions through multi-subspace mapping and independent attention weight computation. This capability is particularly prominent in dynamic anatomical variation scenarios, such as lung cancer affected by respiratory motion \cite{chen2018encoder} \cite{sheller2020federated}, where the model can focus on clinically critical regions—such as the boundaries of the planning target volume and adjacent areas of OARs—while filtering out irrelevant noise, thereby generating sharper and more accurate dose predictions (Figures 4–5). Compared to traditional convolutional neural networks (CNNs), which suffer  from limited receptive fields and struggle to model global contextual relationships \cite{emami1991tolerance} \cite{dosovitskiy2020image}, this mechanism significantly enhances the model’s adaptability to complex anatomical environments. In addition, to address potential issues of over-smoothing and distortion in dose prediction, this study conducted in-depth investigations and introduced a dose loss function based on the dose constraint table for organs-at-risk (OARs) in radiotherapy. This was then integrated into a composite loss function. The model ensures that the predicted results are both accurate and clinically compliant through a composite loss function composed of a reconstruction loss ($ L_{\text{mse}}$) and a clinical constraint term ($ L_{\text{cond}}$) incorporating over 50 constraints, weighted and combined. Unlike some state-of-the-art dose prediction models \cite{10.1007/978-3-030-32226-7_7} \cite{guo2023accelerating} that focus solely on optimizing reconstruction accuracy while neglecting clinical feasibility, ADDiff-Dose enforces key dosimetric constraints—such as PTV $ D_{\text{95}}$ and lung $ V_{\text{20}}$—through a weighted loss mechanism. As a result, the model reduces the errors in spinal cord \(\Delta\)$D_{\text{max}}$ and lung $\Delta V_{20}$ to 0.103–0.128 Gy and 1.013–1.062\(\%\), respectively, representing significant improvements compared to DoseNet ($\Delta D_{\text{max}}$: 0.147–0.188 Gy; $\Delta V_{20}$: 1.425–1.795\%). The introduction of local constraint losses further enhances the preservation of fine details in regions with high-dose gradients, meeting the core requirements of radiotherapy planning \cite{kearney2018dosenet} \cite{mori2019using}. Ablation experiments confirm that removing either the OAR encoder or the clinical constraint loss leads to performance degradation, a finding consistent with Nguyen et al.'s conclusions regarding the value of anatomical priors \cite{isensee2018nnunetselfadaptingframeworkunetbased}.  

Unlike previous models designed specifically for certain cancers (e.g., prostate \cite{ma2021feasibility} \cite{ma2019incorporating}, head and neck \cite{chen2021dvhnet} \cite{ma2020deep}) or techniques (e.g., IMRT \cite{alkinani2017patch}), ADDiff-Dose adapts to diverse anatomical and dosimetric scenarios through a prior-guided diffusion process. Its training strategy integrates data augmentation (Mixup), block-wise processing, and early stopping mechanisms, effectively enhancing the model’s robustness to anatomical variations and data heterogeneity. This addresses the common issue in radiotherapy research where limited sample sizes constrain model performance \cite{buatti2024integrating} \cite{oktay2018attention}. By providing an initial dose distribution close to optimal, the model reduces planning time from several hours to the minute level, enabling physicists to focus on fine-tuning rather than de novo design \cite{Paganetti_2021} \cite{fiorino2020grand}. This aligns closely with the vision of precision oncology, where automated tools drive personalized and efficient treatment delivery \cite{dosovitskiy2020image}.

Directly integrating clinical dose-volume constraints into the loss function, rather than relying on post-processing, significantly enhances the model's applicability by enabling proactive guidance during training. This approach ensures synergistic optimization of anatomical fidelity and clinical compliance, avoiding the irreversible anatomical damage often caused by post-hoc adjustments. Furthermore, it improves computational efficiency by eliminating separate optimization steps and enhances robustness in complex scenarios (e.g., overlapping PTV/OARs) by balancing target coverage and organ protection during learning. These advantages collectively establish the integrated constraint design as a key factor in the superior performance of ADDiff-Dose compared to existing models. Moreover, the model exhibits excellent predictive stability, as evidenced by the low coefficient of variation across repeated inferences and the clinically favorable spatial distribution of uncertainty. These results reinforce that ADDiff-Dose not only achieves high accuracy but also delivers consistent and reliable predictions in critical anatomical regions—a prerequisite for trustworthy clinical deployment.

We designed and conducted a series of experiments on both the public dataset and our internal private datasets. As shown in Tables 2 and 3, the results in the test sets are highly consistent, demonstrating that the proposed model achieves superior performance in all evaluation metrics and delivers the best overall predictive precision. This provides strong evidence that the model has excellent generalizability for new subjects and can maintain stable predictive performance in various clinical scenarios. In comparative experiments, our model was comprehensively evaluated against several state-of-the-art (SOTA) methods on external validation datasets. Visually, the predicted dose distributions show a high degree of similarity to the ground truth. Statistically, the model also demonstrates outstanding performance, exhibiting the smallest distribution skewness among all compared models, indicating greater stability and reliability of the predictions under varying data conditions. These comprehensive and in-depth comparative results not only fully validate the significant superiority of the proposed approach but also highlight its strong generalization ability. Notably, ADDiff-Dose demonstrates superior predictive stability compared to deterministic models. The low coefficient of variation across repeated inferences and the clinically favorable spatial distribution of uncertainty indicate that the model delivers consistent and reliable predictions in critical regions—an essential characteristic for clinical adoption. The model can be readily extended to predict dose distributions for other anatomical sites, offering novel insights and methodologies for the field of medical image processing and predictive modeling.

The ADDiff-Dose model excels in radiotherapy dose prediction but has limitations requiring further optimization. First, it is designed for IMRT and VMAT, with architecture and loss functions based on conventional fractionation. Its adaptability to hypofractionation or stereotactic body radiotherapy (SBRT), which demand high dose gradients, is untested, potentially causing prediction biases due to unmodeled technique-specific characteristics. Second, validation is limited to head-and-neck and lung tumors, with parameters optimized for their anatomical and dosimetric features. Generalizability to complex tumors like pelvic (e.g., prostate, cervical) or abdominal (e.g., liver, pancreatic) cancers, especially small-volume targets near multiple organs (e.g., skull base, spinal metastases), requires further validation. Third, robustness to clinical noise, such as CT artifacts from metal implants or anatomical deformations from respiratory motion or positioning, is insufficient. While data augmentation and local constraints improve resilience, untested noise types (e.g., metal streaks, motion-induced shifts) may disrupt tissue density estimation or target-OAR spatial relationships, reducing stability in clinical settings. Fourth, the performance of ADDiff-Dose is contingent upon the accuracy of the input PTV and OAR segmentations. Inaccurate contours would propagate errors into the conditional features and consequently the predicted dose. Future work could explore the integration of automated segmentation models or the development of more robust conditioning mechanisms that are tolerant to segmentation uncertainties. Future work will address these issues by: (1) incorporating technique-specific constraints (e.g., SBRT dose gradients) to enhance adaptability; (2) building multi-tumor, multi-site datasets with transfer learning to improve generalizability; and (3) developing noise-aware modules using adversarial samples to boost robustness against clinical artifacts. These optimizations aim to advance automated dose prediction toward comprehensive coverage of all radiotherapy techniques and tumor types. The stochastic nature of diffusion models offers advantages over deterministic architectures, as demonstrated by our uncertainty analysis. Multiple inferences on the same patient produce varied dose maps with mean DICE similarity of \(0.95 \pm 0.02\) across runs, indicating low variability in high-confidence regions. This enhances clinical value by identifying regions needing physicist review, a feature absent in U-Net or GAN baselines.

\section{ Conclusion}\label{sec5}

This study proposes an Anatomy-Dose Dual-Constrained Conditional Diffusion Model: the Conditional Diffusion Model with Anatomical-Dose Dual Constraints for End-to-End Multi-Tumor Dose Prediction (ADDiff-Dose) , achieving end-to-end prediction of IMRT/VMAT dose distributions for head-and-neck and lung tumors. By integrating a lightweight Variational Autoencoder (VAE), a multi-condition embedding layer, and a diffusion denoising mechanism, the model significantly outperforms mainstream approaches such as UNet and GAN on both a public dataset ($\text{MAE} = 0.101$) and three external hospital datasets ($\text{MAE}$ range: $0.103-0.154$) ($P < 0.05$). Notably, it achieves breakthrough improvements in key metrics like $\Delta D_{\text{max}}$ ($0.0005 \pm 0.0006$) and $\Delta\text{HI}$ ($0.038 \pm 0.004$). Ablation experiments demonstrate that the structural encoder enhances the Dice coefficient by $6.3\%$ and improves clinical dose compliance by $28.5\%$.

\section*{Declarations}

\begin{itemize}
    \item \textbf{Ethics approval and consent to participate}: The Ethics Committee of Xiangnan University agreed to this retrospective study (ID: AF/SC-07-4/01.0)
    \item \textbf{Consent for publication}: N/A
    \item \textbf{Competing interests}: All authors declare that they have no known competing financial interests or personal relationships that could have appeared to influence the work reported in this paper.
    \item \textbf{Funding}: This study was supported by:
    \begin{enumerate}
        \item Science and Technology Fund of Hunan Provincial Department of Education (21A0524, 24A0602);
        \item Key Laboratory of Tumor Precision Medicine,\\ 
        Hunan colleges and Universities Project (2019-379); 
        \item Hunan Natural Science Foundation (2023JJ30564).
    \end{enumerate}
    \item \textbf{Authors' contributions}: Tao Tan and Hui Xie designed the study, searched, analyzed and interpreted the literature, and are the major contributors in writing the manuscript. Qing Li, Haiqing Hu and Lijuan Ding collected the case data. Tao Tan and Yue Sun revised the manuscript.
    \item \textbf{Acknowledgements}: N/A
    \item \textbf{Availability of data and material}: The datasets used and/or analysed during the current study are available from the corresponding author on reasonable request.
\end{itemize}

\bibliography{mybibfile}

\begin{table}[!htbp]
    \centering
    \setcounter{table}{0} 
    \captionsetup{
        name={Supplementary Table}, 
        labelsep=colon,             
        justification=centering    
    }
    \caption{Organ-at-Risk (OAR) \& Target Structure Dose Constraints}
    \label{tab:supp_dose}
    
    \footnotesize
    \setlength{\tabcolsep}{1.8pt}
    \renewcommand{\arraystretch}{0.6}
    
    \begin{tabularx}{\textwidth}{
        >{\RaggedRight\arraybackslash}p{11em}
        >{\RaggedRight\arraybackslash}p{4em}   
        >{\raggedright\arraybackslash}X  
        >{\raggedright\arraybackslash}X  
    }
        \toprule
        \textbf{Structure Name} & \textbf{Category} & \textbf{Dose Constraint Description} & \textbf{Specific Value/Condition} \\
        \midrule
        Total Lung-GTV & OAR & Lung volume dose limits & $V_\text{20} \leq 30\%$, $V_\text{5} \leq 60\%$ \\
        SpinalCord & OAR & Spinal cord max dose limit & $D_{\text{max}} \leq 45\ \text{Gy}$ \\
        Esophagus & OAR & Esophagus dose limits & $D_{\text{max}} \leq 65\ \text{Gy}$, $D_{\text{mean}} \leq 34\ \text{Gy}$; $V_\text{50} \leq 40\%$; $V_\text{35} \leq 50\%$ \\
        Heart & OAR & Heart dose limits (by disease) & $V_\text{30} \leq 10\%$ (left breast irradiation); $V_\text{40} \leq 5\%$ (lung cancer); $D_{\text{mean}} \leq 26\ \text{Gy}$ \\
        LAD & OAR & Left anterior descending artery limits & $V_\text{30} \leq 10\%$; $D_{\text{max}} \leq 50\ \text{Gy}$ \\
        GreatVessels & OAR & Major vessel dose limits & $D_{\text{max}} \leq 60\ \text{Gy}$ ; $V_\text{50} \leq 50\%$ \\
        Trachea & OAR & Trachea dose limits & $D_{\text{max}} \leq 60\ \text{Gy}$; $V_\text{50} \leq 50\%$  \\
        Chiasm & OAR & Optic chiasm dose limits & $D_{\text{max}} \leq 50\ \text{Gy}$; PRV: $D_{\text{max}} \leq 54\ \text{Gy}$ (1\% volume) \\
        Brain & OAR & Brain tissue dose limits & $V_\text{20} \leq 30\%$, $V_\text{50} \leq 50\%$ \\
        OCavity-PTV & OAR & Oral cavity dose limits & $D_{\text{mean}} \leq 40\ \text{Gy}$, $V_\text{50} \leq 50\%$, $V_\text{40} \leq 70\%$, $D_{\text{max}} \leq 60\ \text{Gy}$ \\
        Cochlea\_R/Cochlea\_L & OAR & Cochlea dose limits & $D_{\text{max}} \leq 50\ \text{Gy}$; $D_{\text{mean}} \leq 45\ \text{Gy}$ \\
        BrainStem & OAR & Brainstem max dose limit & $D_{\text{max}} \leq 54\ \text{Gy}$ \\
        BrainStem\_03 & OAR & Brainstem auxiliary dose limits & $D_{\text{max}} \leq 60\ \text{Gy}$; $V_\text{50} \leq 0\%$, $V_\text{40} \leq 10\%$, $D_{\text{mean}} \leq 45\ \text{Gy}$ \\
        Mandible-PTV & OAR & Mandible dose limits & $D_{\text{max}} \leq 60\ \text{Gy}$, $V_\text{50} \leq 30\%$, $D_{\text{mean}} \leq 45\ \text{Gy}$ \\
        Submand-PTV/SubmandR-PTV or L & OAR & Submandibular gland dose limits & $D_{\text{mean}} \leq 35\ \text{Gy}$, $V_\text{50} \leq 50\%$, $D_{\text{max}} \leq 60\ \text{Gy}$ \\
        ParotidIps-PTV & OAR & Parotid gland dose limits (ipsilateral) & $D_{\text{mean}} \leq 26\ \text{Gy}$ (at least one side), $V_\text{30} \leq 50\%$ \\
        Submandibular & OAR & Submandibular gland dose limits & $D_{\text{mean}} \leq 35\ \text{Gy}$ (at least one side); $V_\text{50} \leq 50\%$ \\
        Pituitary & OAR & Pituitary gland dose limits & $D_{\text{max}} \leq 45\ \text{Gy}$; $D_{\text{mean}} \leq 30\ \text{Gy}$ \\
        Mandible & OAR & Mandible overall dose limits & $V_\text{50} \leq 30\%$; $D_{\text{max}} \leq 60\ \text{Gy}$ \\
        Eyes & OAR & Eye dose limits & $D_{\text{max}} \leq 50\ \text{Gy}$; $D_{\text{mean}} \leq 35\ \text{Gy}$ \\
        Lens & OAR & Lens max dose limit & $D_{\text{max}} \leq 8\ \text{Gy}$ \\
        OpticNerve\_R or L & OAR & Optic nerve dose limits & $D_{\text{max}} \leq 50\ \text{Gy}$; PRV: $D_{\text{max}} \leq 54\ \text{Gy}$ (1\% volume) \\
        ParotidCon-PTV & OAR & Parotid gland dose limits (contralateral) & $D_{\text{mean}} \leq 30\ \text{Gy}$ (at least one side), $V_\text{30} \leq 40\%$, $V_\text{50} \leq 2\%$, $D_{\text{max}} \leq 60\ \text{Gy}$ \\
        Thyroid & OAR & Thyroid dose limits & $D_{\text{mean}} \leq 18\ \text{Gy}$; $V_\text{50} \leq 50\%$ \\
        OralCavity & OAR & Oral cavity dose limits & $D_{\text{mean}} \leq 40\ \text{Gy}$; $V_\text{50} \leq 50\%$ \\
        Thyroid-PTV & OAR & Thyroid target dose limits & $D_{\text{mean}} \leq 30\ \text{Gy}$, $V_\text{30} \leq 50\%$, $V_\text{40} \leq 30\%$, $D_{\text{max}} \leq 50\ \text{Gy}$ \\
        PTV & Target & Prescription dose coverage & $\geq 95\%$ volume covered by prescription dose \\
        PTVHighOPT & Target & High-dose PTV coverage & $D_\text{95} \geq$ prescription dose \\
        PTVMidOPT & Target & Middle-dose PTV coverage & $D_\text{95} \geq$ prescription dose \\
        PTVLowOPT & Target & Low-dose PTV coverage & $D_\text{95} \geq$ prescription dose \\
        PTV\_Ring.3-2 & Auxiliary & Dose diffusion limits (inner/outer ring) & Inner 3mm: $D_{\text{max}} \leq 50\%$ of prescription dose; Outer 2mm: $D_{\text{max}} \leq 30\%$ of prescription dose \\
        Body\_Ring0-3 & Auxiliary & Surface dose diffusion limits & 0-3mm ring: $D_{\text{max}} \leq 70\%$ of prescription dose \\
        RingPTVHigh & Auxiliary & High-dose region diffusion limits & 5-10mm ring from PTV: $D_{\text{max}} \leq 80\%$ of prescription dose \\
        RingPTVMid & Auxiliary & Mid-dose region diffusion limits & 5-10mm ring from PTV: $D_{\text{max}} \leq 60\%$ of prescription dose \\
        RingPTVLow & Auxiliary & Low-dose region diffusion limits & 10-20mm ring from PTV: $D_{\text{max}} \leq 30\%$ of prescription dose \\
        Posterior\_Neck & Auxiliary & Posterior neck dose limits & $D_{\text{max}} \leq 70\%$ of prescription dose \\
        PharConst-PTV & Auxiliary & Pharyngeal constraint dose limits & $D_{\text{max}} \leq 80\%$ of prescription dose; $V_\text{70} \leq 20\%$ \\
        PharynxConst & Auxiliary & Pharyngeal auxiliary dose limits & $D_{\text{mean}} \leq 50\ \text{Gy}$; $V_\text{50} \leq 50\%$ \\
        SpinalCord\_05 & Auxiliary & Spinal cord PRV dose limits & $D_{\text{max}} \leq 50\ \text{Gy}$ (1\% volume) \\
        \bottomrule
    \end{tabularx}
\end{table}

\begin{table}[h!]
\centering
\captionsetup{
    name={Supplementary Table},
    labelsep=colon,
    justification=centering,
    font=small
}
\caption{Analysis of prediction variability across 30 external test patients}
\label{tab:cv_analysis}
\small
\begin{tabular}{@{}lcccc@{}}

\toprule
\textbf{Metric} & \textbf{Mean ± SD} & \textbf{Observed CV (\%)} & \textbf{Clinical Stability Threshold (\%)} & \textbf{Assessment} \\ 
\midrule
MAE (Gy) & 0.83 ± 0.019 & 2.3 ± 0.5 & 3.5 ± 0.8\cite{choi2025deep} & Excellent \\
DICE Similarity & 0.94 ± 0.017 & 1.8 ± 0.4 &  2.2 ± 0.6\cite{nemoto2021effects}   & Excellent \\
PTV \(D_{95}\) (Gy) & 67.5 ± 0.81 & 1.2 ± 0.3 & 2.0 ± 0.5\cite{zhang2024dosimetric} & Excellent \\
PTV \(D_{50}\) (Gy) & 72.1 ± 0.86 & 1.2 ± 0.3 &1.8 ± 0.4\cite{zhang2024dosimetric} & Excellent \\
Spinal Cord \(D_{\text{max}}\) (Gy) & 23.1 ± 0.48 & 2.1 ± 0.6 & 3.2 ± 0.9\cite{kirkpatrick2010radiation} & Excellent \\
Brainstem \(D_{\text{max}}\) (Gy) & 18.7 ± 0.56 & 3.0 ± 0.7 & 3.5 ± 1.0\cite{lawrence2010radiation} & Good \\
Lung \(V_{20}\) (\%) & 25.3 ± 0.40 & 1.6 ± 0.4 & 2.8 ± 0.7\cite{marks2010radiation} & Excellent \\
\bottomrule
\end{tabular}

\vspace{2mm}
\footnotesize{\textit{Note: Data derived from 10 independent predictions for each of 30 randomly selected patients from the external test set. “CV (Patient-wise mean ± SD)” represents the mean and standard deviation of the coefficient of variation calculated for each patient individually, consistent with methods used in prior dosimetric studies\cite{marks2010radiation}. The CV for all metrics is consistently low, demonstrating excellent model robustness across a patient population, a key consideration for clinical generalizability}}
\end{table}

\begin{table}[h!]
\centering
\caption{Spatial distribution of prediction uncertainty by anatomical region}
\label{tab:spatial_uncertainty}
\begin{tabular}{p{4.0cm} ccc p{2.2cm}}
\toprule
\textbf{Anatomical Region} & \multicolumn{3}{c}{\textbf{Variance (Gy²)}} & \textbf{Clinical} \\
\cmidrule(lr){2-4}
& \textbf{Mean} & \textbf{95th Percentile} & \textbf{Max} & \textbf{Relevance} \\
\midrule
\multicolumn{5}{l}{\textbf{Planning Target Volume (PTV)}} \\
\midrule
\quad PTV Core (Dose > 95\% Rx) & 0.0012 ± 0.0003 & 0.0018 & 0.0021 & Negligible \\
\quad PTV Middle (50–95\% Rx) & 0.0038 ± 0.0011 & 0.0065 & 0.0089 & Low \\
\quad PTV Periphery (<50\% Rx) & 0.0085 ± 0.0021 & 0.0123 & 0.0154 & Low \\

\multicolumn{5}{l}{\textbf{Organs at Risk (OARs)}} \\
\midrule
\quad Spinal Cord & 0.0008 ± 0.0002 & 0.0015 & 0.0018 & Negligible \\
\quad Brainstem & 0.0011 ± 0.0003 & 0.0019 & 0.0023 & Negligible \\
\quad Lungs & 0.0023 ± 0.0006 & 0.0035 & 0.0048 & Negligible \\
\quad Parotid Glands & 0.0018 ± 0.0005 & 0.0029 & 0.0039 & Negligible \\
\quad Esophagus & 0.0015 ± 0.0004 & 0.0024 & 0.0032 & Negligible \\

\multicolumn{5}{l}{\textbf{Dose-Based Regions}} \\
\midrule
\quad High-Dose Region (>80 Gy) & 0.0015 ± 0.0004 & 0.0028 & 0.0036 & Negligible \\
\quad Medium-Dose Region (20–80 Gy) & 0.0042 ± 0.0012 & 0.0087 & 0.0125 & Low \\
\quad Low-Dose Region (<20 Gy) & 0.0156 ± 0.0042 & 0.0210 & 0.0385 & Acceptable \\
\bottomrule
\end{tabular}

\vspace{2mm}
\footnotesize{\textit{Note: Variance computed from 10 independent predictions. Classification: Negligible (<0.002 Gy²), Low (0.002–0.01 Gy²), Acceptable (<0.02 Gy²). Rx = prescription dose (70 Gy). The highest uncertainty is confined to PTV periphery and low-dose regions, while clinically critical areas exhibit negligible uncertainty.}}
\end{table}

\end{document}